\documentclass{article}

\usepackage[preprint]{neurips_2026}

\usepackage[utf8]{inputenc} 
\usepackage[T1]{fontenc}    
\usepackage{hyperref}       
\usepackage{url}            
\usepackage{booktabs}       
\usepackage{amsfonts}       
\usepackage{nicefrac}       
\usepackage{microtype}      
\usepackage{xcolor}         
\usepackage{amsmath}        

\newcommand{\dshape}{\bar{d}_{\text{shape}}}   
\newcommand{\dxpt}{d_{\text{xpt}}}             
\usepackage{algorithm}
\usepackage{algpseudocode}
\usepackage{graphicx}
\usepackage{subcaption}
\usepackage{multirow}

\title{Dynamic Plasma Shape Control with Arbitrary Sensor Subsets}

\author{%
  D.~Sorokin$^{1}$\thanks{Corresponding author: \texttt{ds@nextfusion.org}} \quad
  M.~Stokolesov$^{1}$ \quad
  A.~Granovskiy$^{1}$ \quad
  I.~Prokofyev$^{1}$ \quad
  E.~Adishchev$^{1}$ \\
  \textbf{M.~Nurgaliev}$^{1}$ \quad
  \textbf{E.~Khayrutdinov}$^{1}$ \quad
  \textbf{G.~Subbotin}$^{1}$ \quad
  \textbf{R.~Clark}$^{2}$ \quad
  \textbf{D.~Orlov}$^{2}$ \\[0.4em]
  $^{1}$Next Step Fusion, Bertrange, L-8070, Luxembourg \\
  $^{2}$Center for Energy Research, University of California San Diego, CA~92093, USA \\
}

\begin{document}

\maketitle

\begin{abstract}
Plasma shape control in tokamaks requires a real-time controller that tracks dynamically changing shape targets while tolerating diagnostic failures.
Classical approaches decompose the problem into equilibrium reconstruction followed by a linear controller, and assume a fixed, fully operational sensor set.
We present a reinforcement learning agent that addresses both limitations simultaneously.
The agent is trained in NSFsim, a high-fidelity tokamak simulator configured for DIII-D, on a curated dataset of 120 experimental plasma shapes.
The shape targets are resampled as random step changes every 0.25\,s, exposing the agent to diverse transitions across the full shape envelope.
At test time the agent zero-shot tracks dynamic shape sequences; on a held-out static configuration in simulation it achieves a mean shape error of 2.01\,cm, and dynamic trajectory following is demonstrated qualitatively in simulation and on the physical device.
Diagnostic dropout randomly masks 30\% of magnetic sensors per episode, yielding a single policy robust to arbitrary sensor subsets without backup controllers or mode-switching logic.
An asymmetric actor-critic architecture with privileged equilibrium information improves value estimation under partial observability; an auxiliary shape reconstruction head on the actor enables end-to-end shape reconstruction from raw diagnostics and serves as an interpretability tool for policy analysis.
The policy transfers to experimental DIII-D shots, where it directly commands the coil actuators on two dynamic shape maneuvers, and to the independent GSevolve simulator.
\end{abstract}

\section{Introduction}

Precise plasma shape control is a fundamental requirement for safe and efficient tokamak operation.
Plasma boundary geometry affects energy confinement, heat load distribution on plasma-facing components, and plasma stability.
Practical operation further demands that shape targets change dynamically within a single shot: divertor sweeping to spread heat loads, transitions between plasma regimes, and experiment-specific shape sequences all require a controller that can track varying targets in real time.

Classical approaches to plasma shape control at devices such as DIII-D~\cite{687101} and JET~\cite{jet} decompose this problem into two sequential stages.
First, RTEFIT (Real-Time Equilibrium FITting)~\cite{rtefit} estimates the plasma boundary from magnetic diagnostic signals.
Second, a linear multi-input multi-output (MIMO) controller, designed around a linearized plasma response model, issues coil actuator commands to minimize deviations from target shape.
This two-stage pipeline has well-known limitations: reconstruction accuracy degrades when diagnostics fail, linear controllers cannot handle large shape variations, and the full power supply dynamics are often abstracted away, introducing model mismatch.

Reinforcement learning (RL) offers a path to end-to-end controllers that bypass explicit reconstruction and handle nonlinear plasma dynamics directly.
Landmark results at TCV~\cite{Degrave2022MagneticCO} and DIII-D~\cite{subbotin2025reconstruction} demonstrated that RL agents can control tokamaks from raw magnetic observations.
However, these agents assume a fixed, fully operational diagnostic set --- a fragile assumption in practice, where individual probes and loops can fail between shots or even mid-shot.
Prior RL works also target static setpoints or pre-planned equilibrium sequences rather than arbitrary dynamic shape targets.
Finally, end-to-end agents that bypass explicit reconstruction offer no insight into which diagnostics drive control decisions --- a property important for deployment and fault analysis.

We present a single RL agent that addresses all three challenges simultaneously.
The agent is trained in NSFsim~\cite{nsfsim}, a physics-based simulator that models the full DIII-D power system dynamics --- enabling the agent to command the actual chopper hardware.
To cover the operational shape envelope without dense sampling of the 11-dimensional goal space, we curate a dataset of 120 experimental Lower Single Null (LSN) shapes drawn from over 329,000 EFIT equilibria from DIII-D shots (2014--2020).
During training, the target shape is resampled randomly from this dataset every 0.25\,s, exposing the agent to diverse shape transitions across the full envelope and enabling zero-shot generalization to unseen shape sequences at test time.
Robustness to sensor failures is obtained by randomly zeroing 30\% of magnetic probe and flux loop channels per episode (diagnostic dropout), yielding a single policy that operates gracefully under arbitrary diagnostic subsets (validated across random sensor subsets at varying cardinality, Appendix~\ref{app:sensor_importance}) without backup controllers or mode switching.

Our contributions are:
\begin{enumerate}
  \item Zero-shot dynamic shape control: an agent trained on random step changes between 120 experimental LSN configurations generalizes to unseen shape trajectories at inference time, without any trajectory-specific fine-tuning; validated in NSFsim, the independent GSevolve simulator, and experimental DIII-D shots.
  \item Diagnostic dropout: a per-episode input-masking scheme that yields a single policy robust to arbitrary sensor subsets (validated with a DIII-D disabled-sensor mask), without explicit fault detection or controller switching.
  \item An auxiliary shape reconstruction head that acts as a training stabilizer: it reduces $\dshape$ from 4.8 to 4.0\,cm in ablation and improves training stability (episode-length std 21.0 vs.\ 0.7 steps), indicating the auxiliary loss stabilizes learning rather than maximizing mean reward; it also enables gradient-based sensor importance analysis.
\end{enumerate}

\section{Background}

\subsection{Classical plasma shape control}

Plasma shape control in tokamaks has been studied since the early days of large-scale experiments.
The dominant framework is the isoflux algorithm~\cite{isoflux}, which controls the plasma boundary by enforcing equality of magnetic flux at a set of control points on the first wall (the inner surface of the vacuum vessel).
Real-time equilibrium reconstruction codes --- RTEFIT~\cite{rtefit} at DIII-D and LIUQE~\cite{liuqe} at TCV --- convert raw magnetic diagnostic signals into plasma boundary estimates fast enough for closed-loop feedback.
Linear MIMO controllers, derived from normalized coprime factorization~\cite{coprime} or linearized plasma response models~\cite{Mele2019}, then map flux errors to coil actuator commands.
This pipeline has been deployed at DIII-D~\cite{687101}, JET~\cite{jet}, and EAST~\cite{Mele2019} and achieves reliable steady-state shape holding.
More recently, model predictive control (MPC) has been applied at TCV~\cite{tcv_mpc}, removing the linearity constraint while retaining the reconstruction-then-control decomposition.

All classical methods share the first limitation: reconstruction codes were designed for a full sensor set and degrade unpredictably with missing diagnostics, affecting the downstream MIMO controller. Moreover, handling a new failure pattern requires manual weight updates between shots, with no capacity to adapt mid-shot.
Most also share a second: the controller is linearized around a nominal equilibrium, limiting the range of achievable shape changes and making large dynamic variations unreliable.

\subsection{ML-based plasma shape control}

Machine learning has been applied to plasma shape control along two parallel tracks.
The first replaces the equilibrium reconstruction module with a neural surrogate to improve speed and robustness: EFIT-AI~\cite{efitai}, EFIT-NN~\cite{efitnn}, and EFIT-mini~\cite{efitmini} all train networks to approximate the EFIT solver output, retaining the two-stage structure.
\citet{shousha2024} extend this direction by applying dropout training to a kinetic profile reconstruction network, achieving robustness to absent spectroscopic diagnostics (Thomson scattering, charge exchange); the downstream control policy, however, remains unchanged.

The second track applies RL to learn an end-to-end mapping from magnetic observations to actuator commands.
\cite{Degrave2022MagneticCO} trained an MPO~\cite{mpo} agent to control the full plasma configuration at TCV, demonstrating for the first time that RL can operate a tokamak.
Follow-up work applied similar methods to WEST~\cite{10482855,KerbouaBenlarbi2024CurriculumRL} with curriculum learning extensions.
\cite{subbotin2025reconstruction} demonstrated SAC~\cite{sac} with privileged information for real-time shape control at DIII-D.
Hybrid approaches combine ML reconstruction with RL control: \cite{HL3} learns both a dynamics model and a controller using an EFIT-NN surrogate, and \cite{HL3-zeroshot} extends this to zero-shot cross-device transfer.

Despite these advances, existing RL approaches target fixed setpoints or pre-specified sequences rather than arbitrary goal configurations, and assume a fixed, fully operational diagnostic set — none address partial sensor availability.

%

\subsection{Research gap}

To our knowledge, no existing \emph{end-to-end control} method simultaneously addresses: (i) training on a physically realistic distribution of experimental plasma shapes to avoid the curse of dimensionality, (ii) zero-shot generalization to unseen shape trajectories from step-change training, and (iii) robustness to missing magnetic diagnostics in the control policy itself, rather than in the reconstruction stage.

\section{Method}
\label{sec:method}

\begin{figure}[t]
  \centering
  \includegraphics[width=0.7\linewidth]{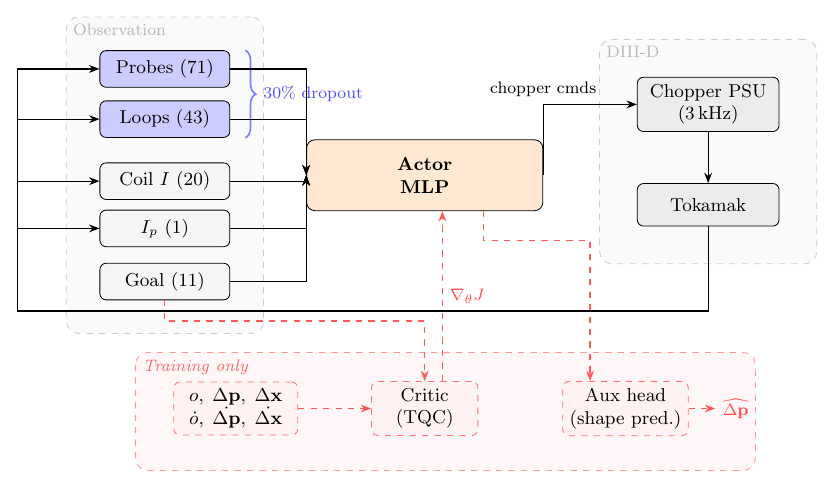}
  \caption{System diagram. The actor MLP receives a 146-dimensional observation $o$ (magnetic probes and flux loops subject to diagnostic dropout, PF coil currents, $I_p$, and shape goal) and outputs chopper commands to the DIII-D power supply. During training only, the asymmetric critic observes the full privileged input $(o,\,\Delta\mathbf{p},\,\Delta\mathbf{x},\,\dot{o},\,\dot{\Delta\mathbf{p}},\,\dot{\Delta\mathbf{x}})$ — the actor's observation augmented with signed pivot-point ($\Delta\mathbf{p}$) and x-point ($\Delta\mathbf{x}$) errors and time derivatives of all inputs — and an auxiliary head predicts the pivot-point error from the actor's penultimate representation.}
  \label{fig:system}
\end{figure}

\subsection{Problem formulation}

We formulate plasma shape control as a Partially Observable Markov Decision Process (POMDP).
The underlying state $s_t$ is the full plasma equilibrium — including the plasma boundary (last closed flux surface, LCFS), kinetic pressure and current profiles — which is not directly observable by the agent.
The actor network receives only partial observations through magnetic diagnostics and coil measurements (Section~\ref{sec:dropout}).
To compensate for partial observability during training, we use an asymmetric actor-critic architecture~\cite{pinto2017asymmetric}: the actor observes only the sensor channels available at PCS, while the critic network additionally has privileged access to plasma boundary, x-point locations, noise-free sensor readings, and time derivatives of all observation components.

\paragraph{Observations.}
The agent observes a 146-dimensional vector composed of: 71 magnetic probe signals, 43 flux loop signals, 20 poloidal field (PF) coil currents, plasma current $I_p$, and an 11-dimensional shape goal $g_t$.
The shape goal encodes the desired plasma configuration as: centroid position $(R_c, Z_c)$, minor radius $a$, upper boundary height $z_{\max}$, upper triangularity $\delta_u$, x-point position $(R_x, Z_x)$, and four squareness values $(\xi_{TI}, \xi_{TO}, \xi_{BI}, \xi_{BO})$ (top/bottom $\times$ inner/outer).
Coil currents, $I_p$, and the shape goal are always fully observed; magnetic probes and flux loops may be masked during training (Section~\ref{sec:dropout}).

\paragraph{Actions.}
The action $a_t$ consists of chopper commands to the DIII-D power supplies.
Rather than commanding coil voltages directly, the agent interfaces with the same actuator representation used in DIII-D operations.
NSFsim~\cite{nsfsim} models the full power system dynamics — chopper circuits, bus voltages, and coil current dynamics — so that the policy learns within the real hardware constraints.
A further DIII-D-specific constraint is the patch panel: multiple coils share a single supply circuit, reducing actuator degrees of freedom and making certain configurations harder to reach.

\paragraph{Reward.}
We measure shape tracking quality via eight 2D pivot points on the LCFS derived from the 11-parameter goal: four structural points (x-point, inner/outer midplane, triangularity apex) and four squareness points that interpolate between adjacent structural pairs toward the bounding-box corners, controlled by $\xi_{TI}, \xi_{TO}, \xi_{BI}, \xi_{BO}$ (see Appendix~\ref{app:pivot_points} for details).
Let $\mathbf{p}_t \in \mathbb{R}^{8 \times 2}$ and $\mathbf{p}^* \in \mathbb{R}^{8 \times 2}$ denote the current and target pivot points, and let $\mathbf{x}_t, \mathbf{x}^* \in \mathbb{R}^2$ denote the current and target x-point positions.
We define two reward components:
\begin{align}
  \dshape &= \frac{1}{8}\sum_{i=1}^{8}\|\mathbf{p}_t^{(i)} - \mathbf{p}^{*(i)}\|_2, \qquad
  \dxpt = \|\mathbf{x}_t - \mathbf{x}^*\|_2, \\
  r_{\text{lcfs}} &= \phi\!\left(\dshape\right), \qquad r_{\text{xpt}} = \phi\!\left(\dxpt\right),
\end{align}
where $\phi(d) = 2\,/\!\left(1 + 19^{\,d/\texttt{geom\_tol}}\right)$, $\texttt{geom\_tol} = 8$\,cm, giving $\phi(0)=1$ (perfect tracking) and $\phi(\texttt{geom\_tol})=0.1$ (poor tracking).
The total reward combines them as a softmax-weighted average:
\begin{equation}
  r_t = \frac{r_{\text{lcfs}}\,e^{\alpha r_{\text{lcfs}}} + r_{\text{xpt}}\,e^{\alpha r_{\text{xpt}}}}{e^{\alpha r_{\text{lcfs}}} + e^{\alpha r_{\text{xpt}}}}, \quad \alpha = -5.
\end{equation}
With $\alpha < 0$ it acts as a soft minimum: the component further from its target receives higher weight, preventing the agent from sacrificing one objective to optimize the other.

\paragraph{Training.}
An episode terminates early if $\dshape$ increases by more than \texttt{geom\_tol}\,=\,8\,cm from its value at the start of the episode, or after 1\,s (1k agent steps).
The agent is trained for $1$M environment steps using TQC (Section~\ref{sec:algorithm}), with target shapes resampled from the experimental dataset every 0.25\,s (Section~\ref{sec:dataset}).

\subsection{Training distribution of goals}
\label{sec:dataset}

The shape goal $g_t$ lies on an 11-dimensional manifold constrained by engineering and physical feasibility requirements.
Uniform random sampling of this manifold is impractical for two reasons: first, it is likely to produce physically unreachable configurations; second, the probability of sampling configurations at the extremes of the operational envelope — maximum elongation, marginal x-point positions, or unusual squareness values — approaches zero under a training budget of $\sim$4,000 target shapes.

We address this by constructing a discrete training distribution from experimental data.
We select 120 plasma shapes from a dataset of over 329,000 equilibria from DIII-D shots between 2014 and 2020 using a greedy diversity criterion: iterating over shots chronologically, a shape is added whenever its mean pivot-point distance to the most recently added shape exceeds 8\,cm, yielding configurations that span the full operational envelope including extreme elongations and x-point locations (Figure~\ref{fig:dataset}).
During each training episode, the target shape $g_t$ is resampled uniformly from this dataset every 0.25\,s, creating a diverse sequence of shape transitions.
The agent must learn to drive the plasma from any shape in the set to any other — sampling a broad distribution of pairwise transitions across the full shape envelope; over $10^6$ training steps the agent encounters approximately 4{,}000 start--target pairs out of $120{\times}120 = 14{,}400$ possible.

\subsection{Diagnostic dropout}
\label{sec:dropout}

Among the 114 maskable diagnostic channels (71 probes + 43 loops), a variable subset may be unavailable at any time due to hardware faults, calibration issues, or deliberate exclusion.
We train the agent to be robust to this variability through \emph{diagnostic dropout}: at the start of each training episode, a binary mask is sampled by independently zeroing each probe and loop channel with probability $p = 0.3$; masking is applied after the standardization block, so a masked channel takes the value zero in standardized space, equivalent to substituting the running mean in raw space.
This mask is held fixed for the entire episode, and surviving channels are rescaled by $1/(1-p)$ to preserve the expected signal magnitude.
The agent receives no explicit mask indicator; it must infer which signals are absent from the pattern of mean-substituted inputs.

Unlike standard neural network dropout, entire input channels corresponding to physical sensors are masked — representing real sensor absence, not stochastic regularization of internal weights.
As we show in Section~\ref{sec:experiments}, the resulting policy approaches the performance of an oracle trained on a fixed sensor mask, while generalizing to arbitrary sensor subsets without requiring advance knowledge of which diagnostics will be available.

\paragraph{Deployment note.}
While training re-samples a new random mask each episode, DIII-D deployment uses a \emph{fixed} mask determined by a pre-shot inspection of unavailable diagnostics, held constant for the entire shot.
The policy is trained at 1\,kHz in NSFsim and deployed at 4\,kHz on PCS and in GSevolve; we observed no degradation from this frequency change.

\subsection{Algorithm: TQC with auxiliary shape loss}
\label{sec:algorithm}

We train the agent using Truncated Quantile Critics (TQC)~\cite{tqc}, a distributional off-policy RL algorithm that reduces overestimation bias by truncating the upper tail of the critic's return distribution.

\paragraph{Asymmetric actor-critic.}
The critic receives the same 146-dimensional observation as the actor, augmented with: (i) privileged information computable from the full simulation state but unavailable on the physical device --- the signed differences between current and target pivot points ($\Delta\mathbf{p} \in \mathbb{R}^{8\times 2}$) and x-point ($\Delta\mathbf{x} \in \mathbb{R}^{2}$); and (ii) time derivatives of all critic inputs, providing explicit velocity information to aid value estimation.

\paragraph{Auxiliary shape reconstruction loss.}
The classical DIII-D Plasma Control System (PCS) pipeline is inherently two-stage: EFIT first reconstructs the plasma shape from magnetic diagnostics, then the controller uses the reconstructed shape to compute actuator commands.
We collapse both stages into a single end-to-end actor network, but preserve the inductive structure of the classical pipeline through an auxiliary loss.
A linear prediction head attached to the actor's penultimate layer is trained to predict the signed difference between the current and target pivot points $\Delta\mathbf{p} \in \mathbb{R}^{8\times 2}$ from the masked diagnostic observations.
The linearity is intentional: success of a linear head implies the penultimate representation has approximately linearized the mapping from raw magnetic observations to shape error — a nontrivial property we confirm empirically in Appendix~\ref{app:aux_reconstruction}, including validation against EFIT on experimental shots.
\begin{equation}
  \mathcal{L}_{\text{aux}} = \|\widehat{\Delta\mathbf{p}} - \Delta\mathbf{p}\|_2^2,
\end{equation}
where $\widehat{\Delta\mathbf{p}}$ is the predicted error and $\Delta\mathbf{p}$ is the ground-truth pivot-point displacement available from simulation.
Weighted equally with the policy gradient objective, this loss provides dense shape-aware supervision at every step, anchoring the actor's internal representation to a physically interpretable geometric quantity throughout training.

The full training procedure, architecture details, and hyperparameters are given in Appendix~\ref{app:algorithm} (Table~\ref{tab:hyperparams}).

\section{Experiments}
\label{sec:experiments}

\paragraph{Environments.}
We use \textbf{NSFsim}~\cite{nsfsim} as the training environment — a free-boundary Grad-Shafranov solver with coupled transport equations that simulates the full DIII-D tokamak including chopper power system dynamics.
For the RL vs.\ isoflux comparison we use \textbf{GSevolve}, an independent tokamak simulator that interfaces directly with the DIII-D Plasma Control System (PCS), allowing both controllers to be run on the same goal trajectories as the experimental DIII-D shots.
Finally, we perform experimental evaluation on the \textbf{DIII-D} tokamak itself via our PCS module.

\paragraph{Baselines.}
We compare against the classical isoflux controller currently deployed on DIII-D, evaluated in GSevolve on the same goal trajectories.
For ablations we train variants of our agent with individual components removed: (i) no auxiliary $\Delta\mathbf{p}$ prediction head, (ii) no asymmetric critic (symmetric critic with no privileged information), and (iii) SAC in place of TQC.
For the diagnostic robustness experiments we additionally compare agents trained with different dropout probabilities $p \in \{0.1, 0.3, 0.5, 0.7\}$ and an oracle policy trained directly on the fixed DIII-D sensor mask, all evaluated on that same mask.

\subsection{NSFsim: control of extreme shapes}

We initialise NSFsim from DIII-D experimental data (coil currents and kinetic profiles from a reference discharge), giving a held-out base shape not seen during training.
From this base we command goal trajectories to four boundary configurations: most compressed ($\min z_{\max}$), most elongated ($\max z_{\max}$), leftmost x-point ($R_{x,\min}$), and rightmost x-point ($R_{x,\max}$).
Table~\ref{tab:NSFsim} reports steady-state metrics averaged over $t > 0.05$\,s.

The agent tracks the base shape with $\dshape = 2.01$\,cm and $\dxpt = 1.69$\,cm despite it being a held-out configuration.
Performance degrades at boundary configurations, as expected: x-point targets show larger $\dxpt$ than $\dshape$ because the power system operates near its voltage limit at the boundaries of the coil current envelope.
The rightmost x-point ($R_{x,\max}$) is the hardest case: large coil current changes are required against the fixed supply voltage, further constrained by the patch panel configuration, producing the highest errors ($\dshape = 5.21$\,cm, $\dxpt = 8.21$\,cm) and lowest reward (0.31).

\begin{table}[h]
\centering
\caption{NSFsim: shape tracking metrics on base and extreme configurations. $\dshape$ and $\dxpt$ are mean distances over the episode; reward is the mean per-step reward averaged over $t > 0.05$\,s to exclude the transition period.}
\label{tab:NSFsim}
\begin{tabular}{lccc}
\toprule
\textbf{Target} & $\dshape$ (cm) & $\dxpt$ (cm) & \textbf{Reward} \\
\midrule
Base shape                    & 2.01 & 1.69 & 0.64 \\
Min $z_{\max}$ (compressed)   & 4.97 & 3.06 & 0.43 \\
Max $z_{\max}$ (elongated)    & 5.13 & 3.43 & 0.35 \\
Min $R_x$ (leftmost x-point)  & 3.52 & 5.34 & 0.36 \\
Max $R_x$ (rightmost x-point) & 5.21 & 8.21 & 0.31 \\
\bottomrule
\end{tabular}
\end{table}

\begin{figure}[t]
  \centering
  \includegraphics[width=\linewidth]{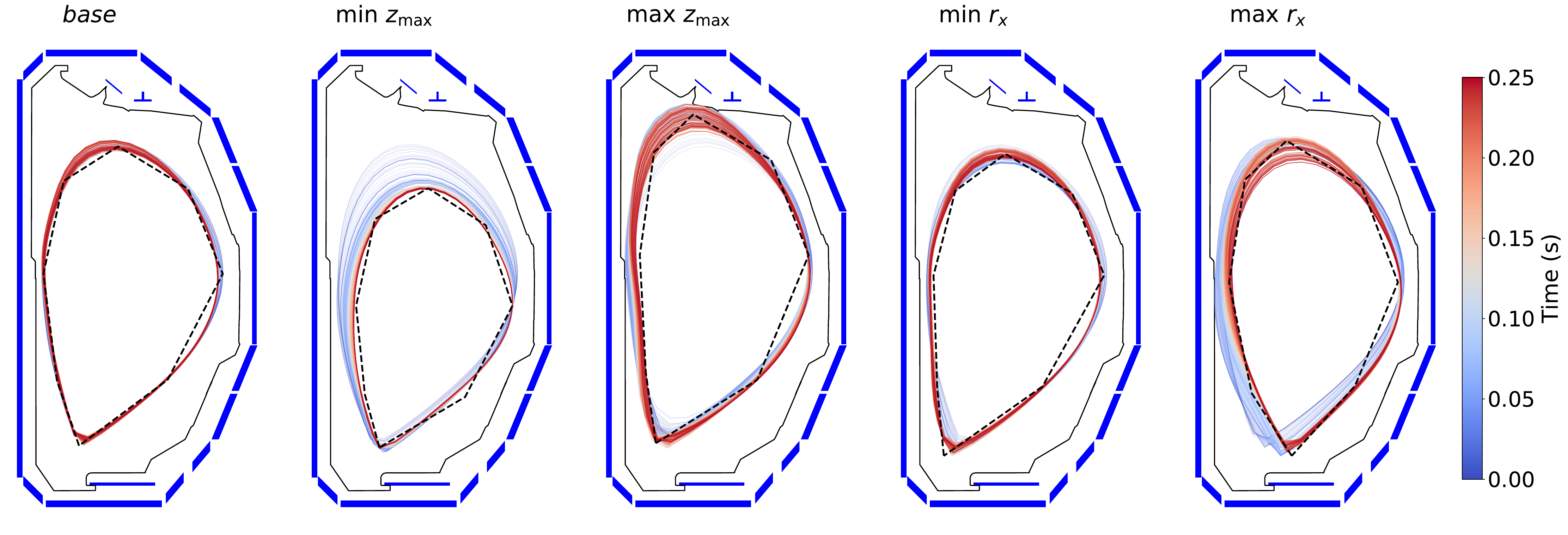}
  \caption{Agent transitions from a common base shape (leftmost panel) to four boundary configurations: minimum upper boundary height ($\min\ z_{\max}$), maximum upper boundary height ($\max\ z_{\max}$), leftmost x-point ($\min\ r_{x}$), and rightmost x-point ($\max\ r_{x}$). Semi-transparent LCFS contours are colored by time (blue $\rightarrow$ red); the target shape is shown as a bold dashed black line.}
  \label{fig:extreme_shapes}
\end{figure}

\subsection{NSFsim: diagnostic robustness}

\begin{figure}[h]
  \centering
  \includegraphics[width=\linewidth]{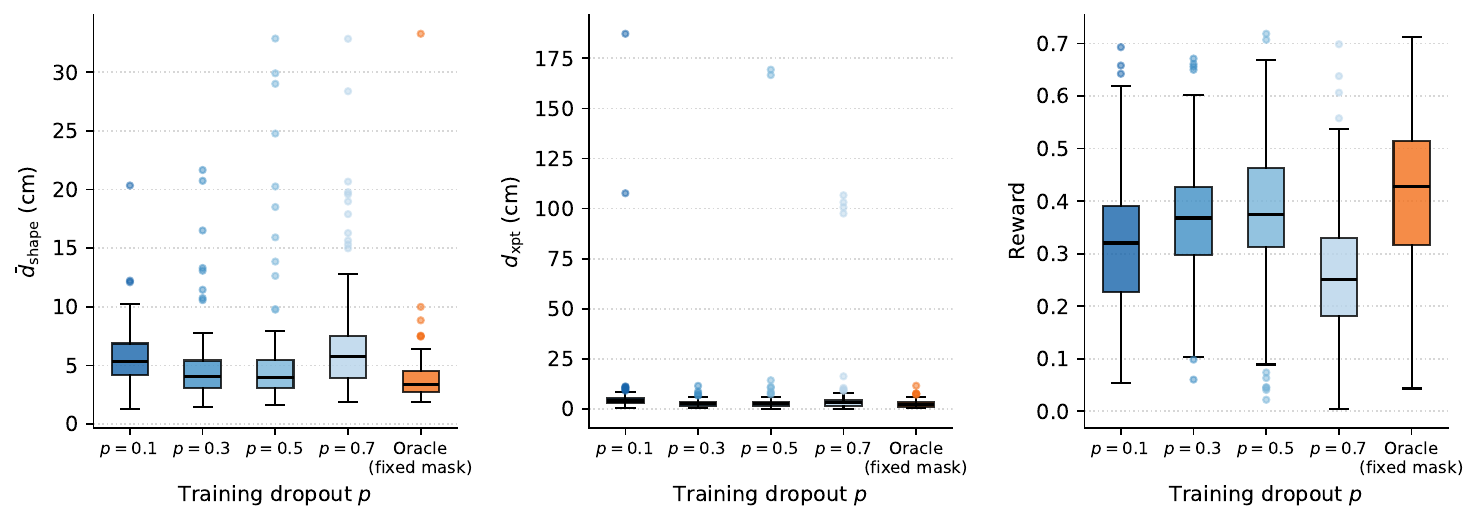}
  \caption{Per-shape distributions of $\dshape$ (left), $\dxpt$ (center), and reward (right) across all 120 NSFsim shapes for agents trained with different dropout probabilities and an oracle trained on the fixed mask, all evaluated on the fixed DIII-D disabled-sensor mask. Boxes show the interquartile range; whiskers extend to $1.5\times$IQR; dots are outliers.}
  \label{fig:dropout_robustness}
\end{figure}

We train agents with dropout probabilities $p \in \{0.1, 0.3, 0.5, 0.7\}$ and evaluate all in NSFsim across all 120 shapes on a fixed sensor mask corresponding to the actual broken diagnostics in DIII-D experimental sessions.
We also include an oracle baseline trained directly on this fixed mask.
Figure~\ref{fig:dropout_robustness} shows the per-shape distributions of $\dshape$, $\dxpt$, and reward for each training condition.

The oracle model achieves the best mean performance ($\dshape = 3.4$\,cm, $\dxpt = 2.0$\,cm), as expected from specialization to this specific mask.
Among dropout-trained agents, $p=0.3$ offers the best tradeoff: mean $\dshape = 4.1$\,cm and $\dxpt = 2.6$\,cm with low per-shape variance ($\sigma_{\dxpt} = 1.6$\,cm).
Despite having a slightly better mean $\dshape$ (3.9\,cm), $p=0.5$ loses the diverted configuration on a subset of shapes --- the plasma transitions from Lower Single Null to a limited configuration --- resulting in large $\dxpt$ deviations ($\sigma_{\dxpt} = 15.4$\,cm); $p=0.7$ shows the same issue ($\sigma_{\dxpt} = 18.2$\,cm).
The $p=0.1$ agent is fragile: trained with a lower failure rate than the real mask, it has not learned to handle ${\approx}30\%$ sensor dropout and degrades accordingly ($\dshape = 5.4$\,cm).
These results confirm that $p=0.3$ is well-matched to the deployment conditions (33 of 114 maskable channels disabled, 28.9\%): it achieves $\dshape = 4.1$\,cm versus $3.4$\,cm for the oracle, a 0.7\,cm gap in exchange for generalization to any sensor subset without prior knowledge of the failure pattern.

\subsection{DIII-D: physical experiments}

\begin{figure}[t]
  \centering
  \begin{minipage}[t]{0.48\linewidth}
    \includegraphics[width=\linewidth]{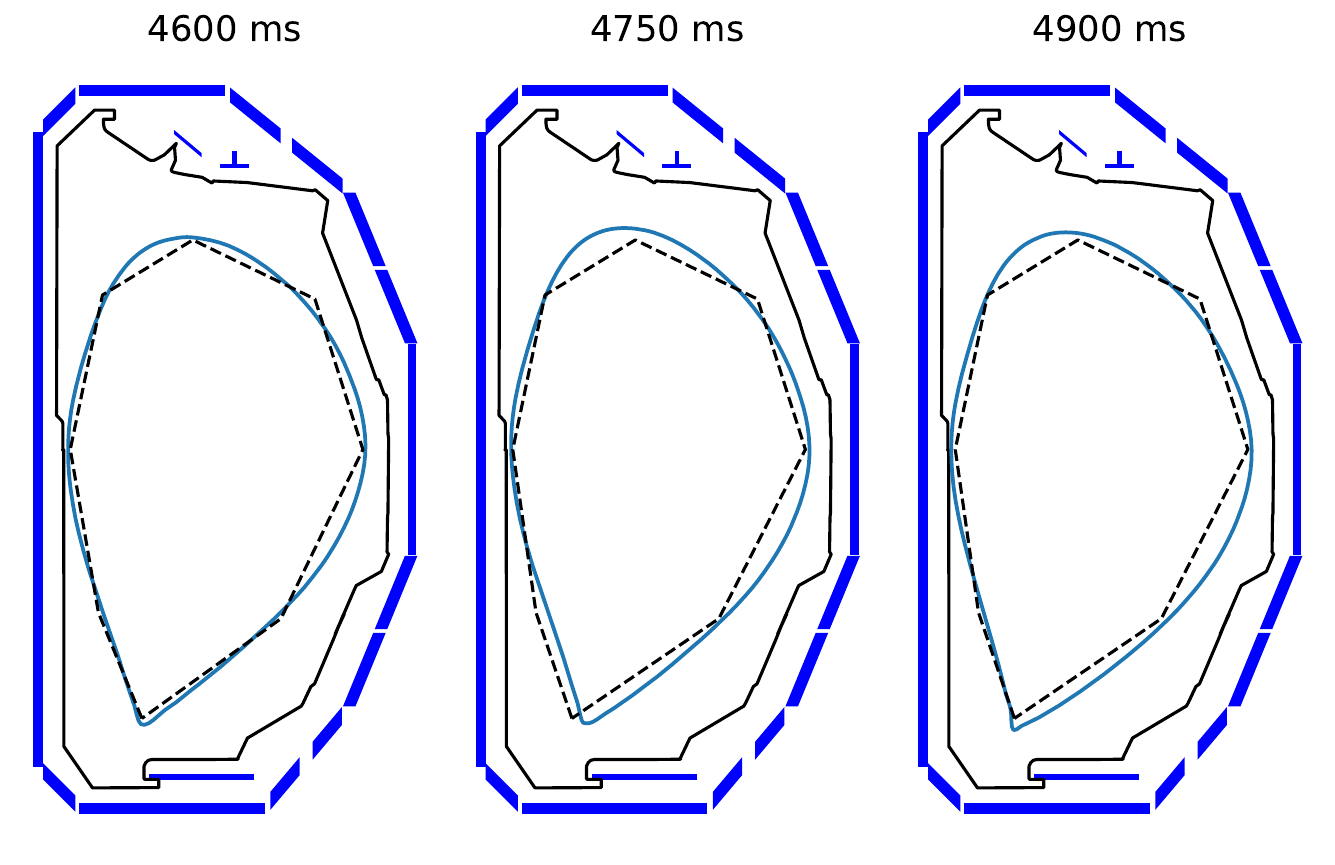}
  \end{minipage}\hfill
  \begin{minipage}[t]{0.48\linewidth}
    \includegraphics[width=\linewidth]{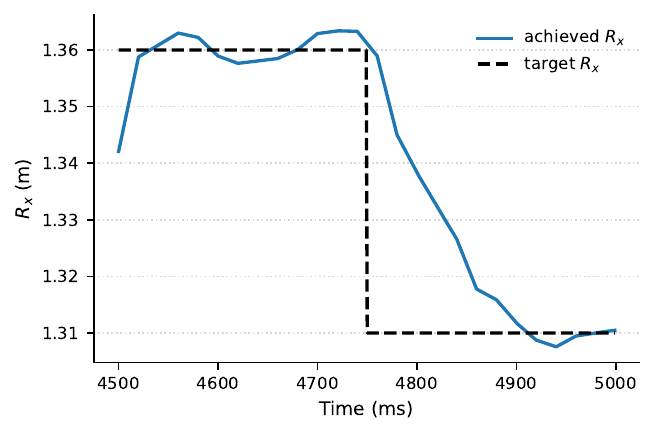}
  \end{minipage}

  \vspace{0.5em}

  \begin{minipage}[t]{0.48\linewidth}
    \includegraphics[width=\linewidth]{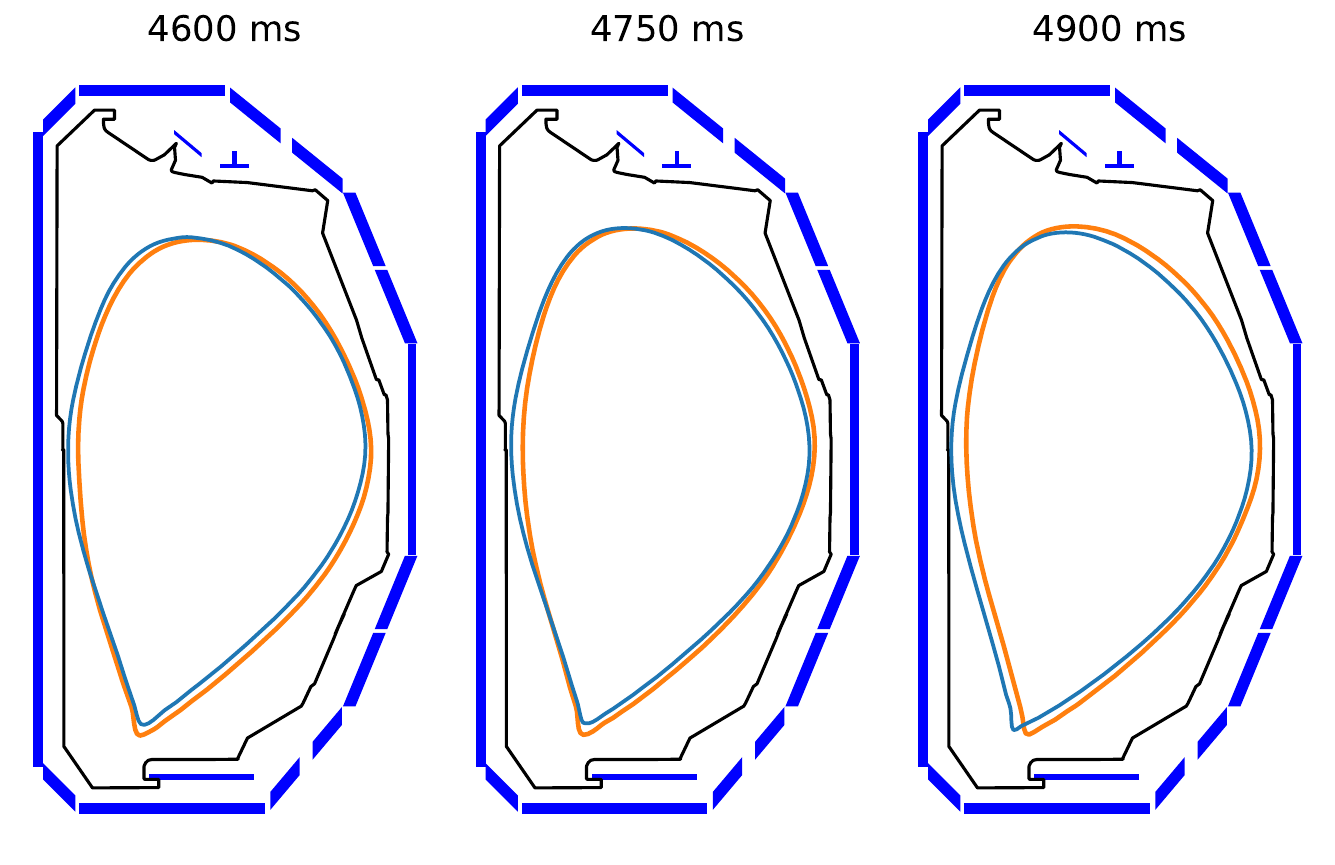}
  \end{minipage}\hfill
  \begin{minipage}[t]{0.48\linewidth}
    \includegraphics[width=\linewidth]{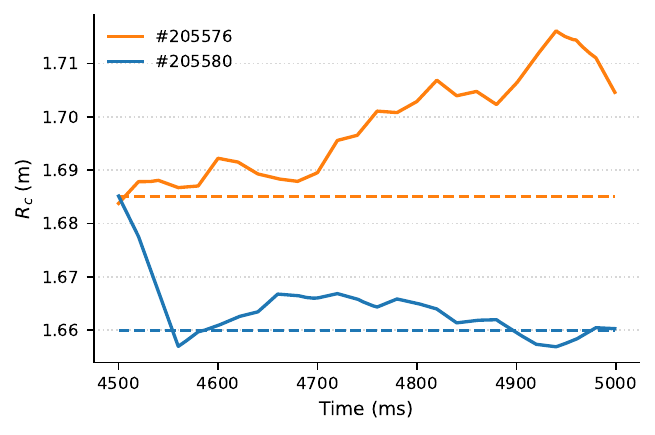}
  \end{minipage}
  \caption{
    Left: tokamak poloidal cross-sections with EFIT-reconstructed plasma boundary — x-point sweep (Discharge \#205580, top) and plasma centroid shift (Discharges \#205580 and \#205576 overlaid, bottom).
    Right: corresponding control metrics — x-point position $R_x(t)$ vs target (top); plasma centroid $R_c(t)$ vs target for both discharges (bottom).
    X-point sweep: $R_x$ moves from 1.36 to 1.31\,m. Centroid shift: $R_c$ changes from 1.685 to 1.660\,m between the two discharges.
  }
  \label{fig:diii-d}
\end{figure}

\begin{figure}[t]
    \begin{minipage}[t]{0.48\linewidth}
    \includegraphics[width=\linewidth]{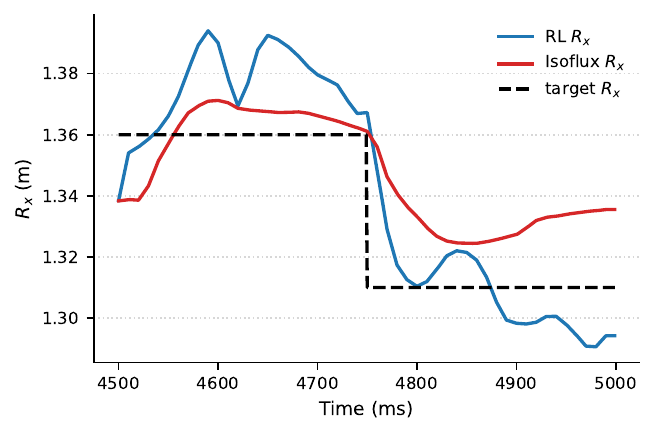}
  \end{minipage}\hfill
  \begin{minipage}[t]{0.48\linewidth}
    \includegraphics[width=\linewidth]{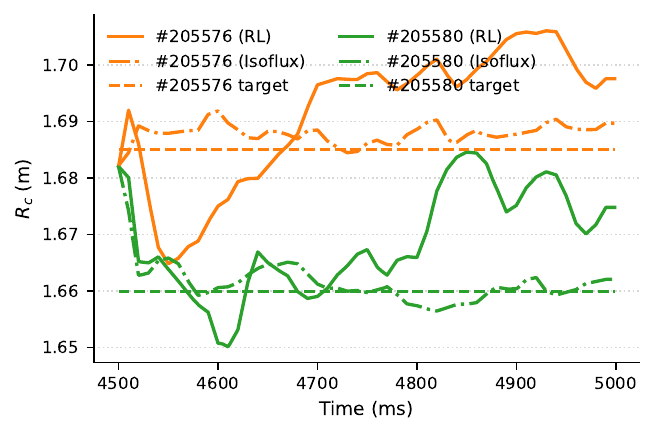}
  \end{minipage}
  \caption{RL vs.\ isoflux comparison in GSevolve on the same goal trajectories as the DIII-D shots.
  \textit{Left} ($R_x$ sweep, Discharge \#205580): blue — RL agent; red — isoflux; dashed black — target.
  \textit{Right} ($R_c$ shift): solid orange — RL (\#205576, $R_c=1.685$\,m); dot-dashed orange — isoflux replay (\#205576); solid green — RL (\#205580, $R_c=1.660$\,m); dot-dashed green — isoflux replay (\#205580); dashed lines — corresponding targets.}
  \label{fig:GSevolve}
\end{figure}

We deployed the trained agent on the DIII-D tokamak in two experiments: an x-point radial sweep (Discharge \#205580) and a plasma centroid shift between two matched discharges (\#205576 and \#205580; note that \#205580 contributes to both experiments).
Figure~\ref{fig:diii-d} shows tokamak poloidal cross-sections with EFIT-reconstructed plasma boundary (left panels) and the corresponding control metrics over time (right panels).
Figure~\ref{fig:GSevolve} provides a quantitative comparison: both the RL agent and the isoflux controller are run in GSevolve on the same goal trajectories as the experimental shots.

\textbf{X-point sweep (Discharge \#205580).}
The x-point radial coordinate $R_x$ was commanded to move from 1.36\,m to 1.31\,m.
The poloidal cross-sections show the plasma boundary shifting as $R_x$ tracks the target; the $R_x(t)$ metric confirms close tracking throughout the maneuver.

\textbf{Plasma centroid shift (Discharges \#205580 and \#205576).}
Two otherwise matched discharges were run with different target centroid positions: $R_c = 1.685$\,m (Discharge \#205576) and $R_c = 1.660$\,m (Discharge \#205580), a 2.5\,cm difference.
The overlaid EFIT boundaries show a clear horizontal separation between the two shots; the $R_c(t)$ metric confirms each target is established and tracked.

Table~\ref{tab:transfer} summarises tracking metrics across NSFsim, GSevolve and DIII-D for both experiments.

The isoflux controller achieves lower steady-state shape error than RL in GSevolve — it was tuned specifically for this operating point — while the RL agent's key advantage is robustness to diagnostic dropout, which isoflux does not provide.
The elevated $\dxpt$ for Discharge \#205576 (6.91 vs 2.76\,cm in GSevolve) is a consistently observed sim-to-real gap, absent in both simulators; one possible explanation is that it reflects a systematic offset in DIII-D raw magnetic readings that EFIT has been calibrated to absorb, leaving isoflux unaffected while shifting the RL policy's inputs out of the training distribution. Both discharges remained safely in LSN throughout, confirming the gap does not affect overall boundary control.
The RL agent in phase~1 of \#205580 achieves lower $\dshape$ on device than in GSevolve (2.58 vs 3.42\,cm), consistent with GSevolve replaying rather than self-consistently evolving kinetic profiles.

\begin{table}[h]
\centering
\caption{Transfer evaluation: shape tracking metrics in NSFsim, GSevolve (RL vs isoflux), and on DIII-D (RL only; isoflux goals differ). Metrics averaged over settled windows ($t > 0.05$\,s after each step change).}
\label{tab:transfer}
\begin{tabular}{lllccc}
\toprule
\textbf{Task} & \textbf{Environment} & \textbf{Controller} & $\dshape$ (cm) & $\dxpt$ (cm) & \textbf{Reward} \\
\midrule
\multirow{4}{*}{\shortstack[l]{\#205576 static hold\\$R_c=1.685$\,m}}
  & NSFsim   & RL (ours) & 4.02 & 1.03 & 0.42 \\
  & GSevolve & RL (ours) & 4.50 & 2.76 & 0.37 \\
  & GSevolve & Isoflux   & 1.87 & 1.04 & 0.71 \\
  & DIII-D   & RL (ours) & 3.41 & 6.91 & 0.20 \\
\midrule
\multirow{4}{*}{\shortstack[l]{\#205580 $R_x$ phase 1\\$R_x=1.36$\,m, $R_c=1.660$\,m}}
  & NSFsim   & RL (ours) & 4.19 & 1.47 & 0.41 \\
  & GSevolve & RL (ours) & 3.42 & 2.50 & 0.47 \\
  & GSevolve & Isoflux   & 2.73 & 1.42 & 0.58 \\
  & DIII-D   & RL (ours) & 2.58 & 2.80 & 0.54 \\
\midrule
\multirow{4}{*}{\shortstack[l]{\#205580 $R_x$ phase 2\\$R_x=1.31$\,m, $R_c=1.660$\,m}}
  & NSFsim   & RL (ours) & 5.59 & 2.08 & 0.28 \\
  & GSevolve & RL (ours) & 3.73 & 2.63 & 0.44 \\
  & GSevolve & Isoflux   & 2.65 & 2.76 & 0.54 \\
  & DIII-D   & RL (ours) & 3.16 & 4.15 & 0.40 \\
\bottomrule
\end{tabular}
\end{table}

\subsection{Ablations}

To isolate each component's contribution from dropout stochasticity, ablation variants are trained without diagnostic dropout and evaluated on the fixed DIII-D disabled-sensor mask across all 120 NSFsim shapes (Appendix~\ref{app:ablations}, Table~\ref{tab:ablations} and Figure~\ref{fig:training_curves}).
Removing the asymmetric critic has the largest effect (reward 0.41→0.37, $\dxpt$ 2.5→3.4\,cm).
Removing the auxiliary loss leaves mean reward unchanged (0.415 vs.\ 0.414, $\ll\sigma{=}0.138$) but raises $\dshape$ from 4.0 to 4.8\,cm and episode-length std from 0.7 to 21.0 steps, indicating frequent early terminations that do not register as average-reward degradation — the loss acts as a training stabilizer, not a reward maximizer.
SAC ($r{=}0.36$, $\dxpt$ std 16.7 vs.\ 1.8\,cm for TQC) occasionally loses x-point control on harder shapes (Figure~\ref{fig:ablation_boxes}).
The auxiliary head enables gradient-based sensor importance analysis (Appendix~\ref{app:sensor_importance}): high-importance magnetic sensors cluster spatially near the pivot points and the inner limiter wall, consistent with where the plasma--wall gap carries the most positional information.
Importance rankings are stable across all trained dropout rates (Spearman $\rho > 0.96$), confirming the structure reflects task geometry rather than training noise.
The auxiliary head also functions as a standalone shape reconstruction module: validated against offline EFIT on experimental DIII-D shots, it achieves $1.21$--$1.43$\,cm mean pivot-point error while running at 4\,kHz on a partially masked sensor set (Appendix~\ref{app:aux_reconstruction}, Figure~\ref{fig:efit_vs_aux}).

\section{Conclusion}

We presented an RL agent for dynamic plasma shape control at DIII-D that simultaneously addresses three open challenges: training-distribution coverage, dynamic trajectory generalization, and variable-diagnostic robustness.
A curated dataset of 120 experimental LSN shapes with random step-change training avoids the curse of dimensionality, and diagnostic dropout produces a single policy robust to arbitrary subsets of the 71 probes and 43 flux loops without backup controllers or mode-switching logic.
An auxiliary shape reconstruction loss, motivated by the classical two-stage EFIT--PCS pipeline, encourages shape-aware representations in the actor; the asymmetric TQC uses privileged information to accelerate training.
The policy transfers from NSFsim to GSevolve and to DIII-D, commanding the actual chopper power supplies on two dynamic shape maneuvers.
Limitations include reliance on a single tokamak geometry and mean substitution for absent sensors; future work targets multi-machine transfer, adaptive dropout scheduling, and dynamic mid-shot masking.

\begin{ack}
Disclaimer: This report was prepared as an account of work sponsored by an agency of the United States Government. Neither the United States Government nor any agency thereof, nor any of their employees, makes any warranty, express or implied, or assumes any legal liability or responsibility for the accuracy, completeness, or usefulness of any information, apparatus, product, or process disclosed, or represents that its use would not infringe privately owned rights. Reference herein to any specific commercial product, process, or service by trade name, trademark, manufacturer, or otherwise does not necessarily constitute or imply its endorsement, recommendation, or favoring by the United States Government or any agency thereof. The views and opinions of authors expressed herein do not necessarily state or reflect those of the United States Government or any agency thereof.

This material is based upon work supported by the U.S. Department of Energy, Office of Science, Office of Fusion Energy Sciences, using the DIII-D National Fusion Facility, a DOE Office of Science user facility, under Awards DE-FC02-04ER54698, DE-FG02-05ER54809, and Next Step Fusion S.a.r.l. 
\end{ack}

\bibliographystyle{unsrtnat}
\bibliography{neurips_2026}

@article{Degrave2022MagneticCO,
  title={Magnetic control of tokamak plasmas through deep reinforcement learning},
  author={Jonas Degrave and Federico Felici and Jonas Buchli and Michael Neunert and Brendan D. Tracey and Francesco Carpanese and Timo Ewalds and Roland Hafner and Abbas Abdolmaleki and Diego de Las Casas and Craig Donner and Leslie Fritz and Cristian Galperti and Andrea Huber and James Keeling and Maria Tsimpoukelli and Jackie Kay and Antoine Merle and J-M. Moret and Seb Noury and Federico Pesamosca and D. Pfau and Olivier Sauter and Cristian Sommariva and Stefano Coda and B. Duval and Ambrogio Fasoli and Pushmeet Kohli and Koray Kavukcuoglu and Demis Hassabis and Martin A. Riedmiller},
  journal={Nature},
  year={2022},
  volume={602},
  pages={414 - 419},
  url={https://api.semanticscholar.org/CorpusID:246904229}
}

@article{subbotin2025reconstruction,
  title = {Demonstration of reconstruction-free static magnetic control of DIII-D plasma with deep reinforcement learning},
  volume = {66},
  ISSN = {1741-4326},
  url = {http://dx.doi.org/10.1088/1741-4326/ae34c6},
  DOI = {10.1088/1741-4326/ae34c6},
  number = {2},
  journal = {Nuclear Fusion},
  publisher = {IOP Publishing},
  author = {Subbotin,  G.F. and Sorokin,  D.I. and Nurgaliev,  M.R. and Granovskiy,  A.A. and Kharitonov,  I.P. and Adishchev,  E.V. and Khairutdinov,  E.N. and Clark,  R. and Shen,  H. and Choi,  W. and Barr,  J. and Orlov,  D.M.},
  year = {2026},
  month = jan,
  pages = {026040}
}

@ARTICLE{10482855,
  author={Kerboua-Benlarbi, S. and Nouailletas, R. and Faugeras, B. and Nardon, E. and Moreau, P.},
  journal={IEEE Transactions on Plasma Science}, 
  title={Magnetic Control of WEST Plasmas Through Deep Reinforcement Learning}, 
  year={2024},
  volume={52},
  number={9},
  pages={3698-3703},
  doi={10.1109/TPS.2024.3377811}}

@inproceedings{KerbouaBenlarbi2024CurriculumRL,
  title={Curriculum Reinforcement Learning for Tokamak Control},
  author={S. Kerboua-Benlarbi and R{\'e}my Nouailletas and Blaise Faugeras and Philippe Moreau},
  booktitle={AI4Research/DemocrAI@IJCAI},
  year={2024},
  url={https://api.semanticscholar.org/CorpusID:274598363}
}

@INPROCEEDINGS{687101,
  author={Walker, M.L. and Humphreys, D.A. and Ferron, J.R.},
  booktitle={17th IEEE/NPSS Symposium Fusion Engineering (Cat. No.97CH36131)}, 
  title={Multivariable shape control development on the DIII-D tokamak}, 
  year={1997},
  volume={1},
  number={},
  pages={556-559 vol.1},
  doi={10.1109/FUSION.1997.687101}}

@article{rtefit,
    doi = {10.1088/0029-5515/38/7/308},
    url = {https://doi.org/10.1088/0029-5515/38/7/308},
    year = {1998},
    month = {jul},
    publisher = {},
    volume = {38},
    number = {7},
    pages = {1055},
    author = {J.R. Ferron and M.L. Walker and L.L. Lao and H.E. St. John and D.A. Humphreys and J.A. Leuer},
    title = {Real time equilibrium reconstruction 
    for tokamak discharge control},
    journal = {Nuclear Fusion}
}

@article{liuqe,
title = {Tokamak equilibrium reconstruction code LIUQE and its real time implementation},
journal = {Fusion Engineering and Design},
volume = {91},
pages = {1-15},
year = {2015},
issn = {0920-3796},
doi = {https://doi.org/10.1016/j.fusengdes.2014.09.019},
url = {https://www.sciencedirect.com/science/article/pii/S0920379614005973},
author = {J.-M. Moret and B.P. Duval and H.B. Le and S. Coda and F. Felici and H. Reimerdes}
}

@article{jet,
  title = {Plasma shape control for the JET tokamak: an optimal output regulation approach},
  author = {M. Ariola and A. Pironti},
  volume = {25},
  ISSN = {1941-000X},
  url = {http://dx.doi.org/10.1109/MCS.2005.1512796},
  DOI = {10.1109/mcs.2005.1512796},
  number = {5},
  journal = {IEEE Control Systems},
  publisher = {Institute of Electrical and Electronics Engineers (IEEE)},
  year = {2005},
  month = oct,
  pages = {65–75}
}

@article{isoflux,
doi = {10.1088/0029-5515/30/10/003},
url = {https://doi.org/10.1088/0029-5515/30/10/003},
year = {1990},
month = {oct},
publisher = {},
volume = {30},
number = {10},
pages = {2013},
author = {Hofmann, F. and Jardin, S.C.},
title = {Plasma shape and position control in highly elongated tokamaks},
journal = {Nuclear Fusion}
}

@ARTICLE{coprime,
  author={Glover, K. and McFarlane, D.},
  journal={IEEE Transactions on Automatic Control}, 
  title={Robust stabilization of normalized coprime factor plant descriptions with H/sub infinity /-bounded uncertainty}, 
  year={1989},
  volume={34},
  number={8},
  pages={821-830},
  doi={10.1109/9.29424}}

@book{efitai,
  title = {EFIT‐AI: Machine Learning and Artificial Intelligence Assisted Equilibrium Reconstruction for Tokamak Experiments and Burning Plasmas (Final Report)},
  url = {http://dx.doi.org/10.2172/2484189},
  DOI = {10.2172/2484189},
  institution = {Office of Scientific and Technical Information (OSTI)},
  author = {Kruger,  Scott and Howell,  Eric},
  year = {2024},
  month = dec 
}

@article{efitnn,
  title={Real-time equilibrium reconstruction by neural network based on HL-3 tokamak},
  author={Zheng, Guohui and Liu, Songfen and Yang, Zongyu and Ma, Rui and Gong, Xinwen and Wang, Ao and Wang, Shuo and Zhong, Wulyu},
  journal={arXiv preprint arXiv:2405.11221},
  year={2024}
}

@article{efitmini,
  title={EFIT-mini: an embedded, multi-task neural network-driven equilibrium inversion algorithm},
  author={Zheng, GH and Liu, SF and Xie, HS and Gu, X and Chen, ZY and Lun, XC and Liu, Y and Li, J and Guo, D and Tao, RY and others},
  journal={Nuclear Fusion},
  volume={65},
  number={10},
  pages={106008},
  year={2025},
  publisher={IOP Publishing}
}

@inproceedings{tcv_mpc,
  title={First experimental demonstration of plasma shape control in a tokamak through Model Predictive Control},
  author={Mele, Adriano and Topalova, Maria A and Galperti, Cristian and Coda, Stefano},
  booktitle={2025 IEEE Conference on Control Technology and Applications (CCTA)},
  pages={820--825},
  year={2025},
  organization={IEEE}
}

@article{mpo,
  title={Maximum a posteriori policy optimisation},
  author={Abdolmaleki, Abbas and Springenberg, Jost Tobias and Tassa, Yuval and Munos, Remi and Heess, Nicolas and Riedmiller, Martin},
  journal={arXiv preprint arXiv:1806.06920},
  year={2018}
}

@inproceedings{sac,
  title={Soft actor-critic: Off-policy maximum entropy deep reinforcement learning with a stochastic actor},
  author={Haarnoja, Tuomas and Zhou, Aurick and Abbeel, Pieter and Levine, Sergey},
  booktitle={International conference on machine learning},
  pages={1861--1870},
  year={2018},
  organization={Pmlr}
}

@inproceedings{HL3,
  title={High-Fidelity Data-Driven Dynamics Model for Reinforcement Learning-based Magnetic Control in HL-3 Tokamak},
  author={Niannian Wu and Zongyu Yang and Rongpeng Li and Ning Wei and Yihang Chen and Qianyun Dong and Jiyuan Li and Guohui Zheng and Xinwen Gong and Feng Gao and Bo Li and Min Xu and Zhifeng Zhao and Wulyu Zhong},
  url={https://api.semanticscholar.org/CorpusID:272689630},
  year={2025}
}

@article{HL3-zeroshot,
  title={Plasma Shape Control via Zero-shot Generative Reinforcement Learning},
  author={Wu, Niannian and Li, Rongpeng and Yang, Zongyu and Xiao, Yong and Wei, Ning and Chen, Yihang and Li, Bo and Zhao, Zhifeng and Zhong, Wulyu},
  journal={arXiv preprint arXiv:2510.17531},
  year={2025}
}

@ARTICLE{nsfsim,
  title     = "Validation of {NSFsim} as a Grad-Shafranov equilibrium solver at
               {DIII}-{D}",
  author    = "Clark, Randall and Nurgaliev, Maxim and Khairutdinov, Eduard and
               Subbotin, Georgy and Welander, Anders and Orlov, Dmitri M",
  journal   = "Fusion Eng. Des.",
  publisher = "Elsevier BV",
  volume    =  211,
  number    =  114765,
  pages     =  114765,
  month     =  "1~" # feb,
  year      =  2025,
  url       = "http://dx.doi.org/10.1016/j.fusengdes.2024.114765",
  doi       = "10.1016/j.fusengdes.2024.114765",
  issn      = "0920-3796,1873-7196",
  language  = "en"
}

@InProceedings{TQC,
  title = 	 {Controlling Overestimation Bias with Truncated Mixture of Continuous Distributional Quantile Critics},
  author =       {Kuznetsov, Arsenii and Shvechikov, Pavel and Grishin, Alexander and Vetrov, Dmitry},
  booktitle = 	 {Proceedings of the 37th International Conference on Machine Learning},
  pages = 	 {5556--5566},
  year = 	 {2020},
  editor = 	 {III, Hal Daumé and Singh, Aarti},
  volume = 	 {119},
  series = 	 {Proceedings of Machine Learning Research},
  month = 	 {13--18 Jul},
  publisher =    {PMLR},
  url = 	 {https://proceedings.mlr.press/v119/kuznetsov20a.html}
}

@article{pinto2017asymmetric,
  title={Asymmetric actor critic for image-based robot learning},
  author={Pinto, Lerrel and Andrychowicz, Marcin and Welinder, Peter and Zaremba, Wojciech and Abbeel, Pieter},
  journal={arXiv preprint arXiv:1710.06542},
  year={2017}
}

@article{Mele2019,
  title = {MIMO shape control at the EAST tokamak: Simulations and experiments},
  volume = {146},
  ISSN = {0920-3796},
  url = {http://dx.doi.org/10.1016/j.fusengdes.2019.02.058},
  DOI = {10.1016/j.fusengdes.2019.02.058},
  journal = {Fusion Engineering and Design},
  publisher = {Elsevier BV},
  author = {Mele,  A. and Albanese,  R. and Ambrosino,  R. and Castaldo,  A. and De Tommasi,  G. and Luo,  Z.P. and Pironti,  A. and Yuan,  Q.P. and Yuehang,  W. and Xiao,  B.J.},
  year = {2019},
  month = Sept,
  pages = {1282–1285}
}

@article{shousha2024,
  title = {Machine learning-based real-time kinetic profile reconstruction in DIII-D},
  volume = {64},
  ISSN = {1741-4326},
  url = {http://dx.doi.org/10.1088/1741-4326/ad142f},
  DOI = {10.1088/1741-4326/ad142f},
  number = {2},
  journal = {Nuclear Fusion},
  publisher = {IOP Publishing},
  author = {Shousha,  Ricardo and Seo,  Jaemin and Erickson,  Keith and Xing,  Zichuan and Kim,  SangKyeun and Abbate,  Joseph and Kolemen,  Egemen},
  year = {2023},
  month = Dec,
  pages = {026006}
}


\newpage
\appendix

\section{Pivot Point Parameterization}
\label{app:pivot_points}

Given the 11-parameter goal $(R_c, Z_c, a, z_{\max}, \delta_u, R_x, Z_x, \xi_{TI}, \xi_{TO}, \xi_{BI}, \xi_{BO})$, define $r_{\min} = R_c - a$, $r_{\max} = R_c + a$, and bounding-box corners $\mathbf{TI}{=}(r_{\min},z_{\max})$, $\mathbf{TO}{=}(r_{\max},z_{\max})$, $\mathbf{BI}{=}(r_{\min},Z_x)$, $\mathbf{BO}{=}(r_{\max},Z_x)$.
The eight pivot points are:
\begin{align*}
\mathbf{p}_1 &= (R_x,\; Z_x) & &\text{x-point} \\
\mathbf{p}_2 &= \tfrac{\mathbf{p}_1+\mathbf{p}_3}{2} + \tfrac{\mathbf{BI}-(R_x,\,Z_c)}{2}\,\xi_{BI} & &\text{bottom-inner squareness} \\
\mathbf{p}_3 &= (r_{\min},\; Z_c) & &\text{inner midplane} \\
\mathbf{p}_4 &= \tfrac{\mathbf{p}_5+\mathbf{p}_3}{2} + \tfrac{\mathbf{TI}-(p_{5,r},\,r_{\min,z})}{2}\,\xi_{TI} & &\text{top-inner squareness} \\
\mathbf{p}_5 &= (R_c - a\,\delta_u,\; z_{\max}) & &\text{topmost point} \\
\mathbf{p}_6 &= \tfrac{\mathbf{p}_5+\mathbf{p}_7}{2} + \tfrac{\mathbf{TO}-(p_{5,r},\,r_{\max,z})}{2}\,\xi_{TO} & &\text{top-outer squareness} \\
\mathbf{p}_7 &= (r_{\max},\; Z_c) & &\text{outer midplane} \\
\mathbf{p}_8 &= \tfrac{\mathbf{p}_1+\mathbf{p}_7}{2} + \tfrac{\mathbf{BO}-(R_x,\,Z_c)}{2}\,\xi_{BO} & &\text{bottom-outer squareness}
\end{align*}
where $p_{5,r} = R_c - a\,\delta_u$ and $r_{\min,z} = r_{\max,z} = Z_c$ denote the $r$- and $z$-coordinates used for axis alignment in the squareness offsets.
A squareness parameter $\xi = 0$ places the point at the midpoint of its two neighboring structural points; $\xi = \pm 1$ displaces it fully to the corresponding bounding-box corner.

\section{Architecture, Algorithm, and Hyperparameters}
\label{app:algorithm}

The actor is a multi-layer perceptron (MLP) that takes the 146-dimensional observation (with dropout-masked probe and loop channels) as input and outputs chopper commands.
The TQC critics are an ensemble of distributional networks that receive the actor's observation concatenated with privileged equilibrium information during training.
The auxiliary shape prediction head is a single linear layer applied to the actor's penultimate representation.
All inputs to the actor and critic are standardized online using running estimates of the mean and standard deviation, updated throughout training.
To improve sim-to-real robustness, independent Gaussian observation noise is added to magnetic probes ($\sigma = 0.01$\,mT), flux loops ($\sigma = 0.01$\,mWb), and coil currents ($\sigma = 100$\,A); power supply voltages are additionally randomized with uniform noise ($\pm50$\,V) as dynamics randomization; see Table~\ref{tab:hyperparams}.

\begin{algorithm}[h]
\caption{Training: TQC with asymmetric critic and auxiliary shape loss}
\label{alg:training}
\begin{algorithmic}[1]
\State \textbf{Input:} NSFsim environment, shape dataset $\mathcal{D}$ (120 LSN shapes)
\State Initialize actor $\pi_\theta$, critics $\{Q_{\phi_i}\}$, auxiliary head $h_\psi$, replay buffer $\mathcal{B}$
\For{each episode}
  \State Sample masks independently: $m_{\text{probes}} \sim \text{Bernoulli}(1-p)^{71}$, $m_{\text{loops}} \sim \text{Bernoulli}(1-p)^{43}$, hold fixed for episode  \Comment{diagnostic dropout}
  \State Sample initial shape $g_0 \sim \mathcal{D}$; reset NSFsim
  \For{each step $t$}
    \State Observe $o_t$; standardize using running mean $\hat{\mu}$ and std $\hat{\sigma}$; update $\hat{\mu}, \hat{\sigma}$
    \State Apply mask $m$: zero masked channels in standardized space (equivalent to mean substitution in raw space); rescale surviving channels by $1/(1-p)$
    \State $a_t \leftarrow \pi_\theta(o_t)$ \Comment{chopper commands}
    \State Execute $a_t$; receive $r_t$, next state
    \If{$t \bmod 250 = 0$} sample new target $g_t \sim \mathcal{D}$ \EndIf  \Comment{every 0.25\,s}
    \State Store $(o_t, a_t, r_t, o_{t+1}, \Delta\mathbf{p}_t, \Delta\mathbf{x}_t)$ in $\mathcal{B}$
  \EndFor
\EndFor
\For{each gradient update}
  \State Sample minibatch from $\mathcal{B}$
  \State Compute critic input: $[o_t,\, \Delta\mathbf{p}_t,\, \Delta\mathbf{x}_t,\, \dot{o}_t,\, \dot{\Delta\mathbf{p}}_t,\, \dot{\Delta\mathbf{x}}_t]$; standardize with $\hat{\mu}, \hat{\sigma}$
  \State Update critics $\{Q_{\phi_i}\}$ via TQC loss $\mathcal{L}_{\text{TQC}}$
  \State Update actor $\pi_\theta$ via policy gradient $\nabla_\theta J$
  \State Update auxiliary head $h_\psi$: $\mathcal{L}_{\text{aux}} = \|\widehat{\Delta\mathbf{p}} - \Delta\mathbf{p}\|_2^2$
  \State Total actor loss: $\mathcal{L} = \mathcal{L}_{\text{actor}} + \mathcal{L}_{\text{aux}}$
\EndFor
\State \textbf{Deploy:} actor $\pi_\theta$ only (no critic, auxiliary head retained for analysis)
\end{algorithmic}
\end{algorithm}

\begin{table}[h]
\centering
\caption{Training hyperparameters.}
\label{tab:hyperparams}
\setlength{\tabcolsep}{4pt}
\begin{tabular}{@{}lc@{\hspace{2em}}lc@{}}
\toprule
\textbf{Hyperparameter} & \textbf{Value} & \textbf{Hyperparameter} & \textbf{Value} \\
\midrule
\multicolumn{2}{@{}l}{\textit{Algorithm}} & \multicolumn{2}{l@{}}{\textit{Training}} \\
RL algorithm          & TQC                  & Batch size          & 1024 \\
Critics               & 3                    & Replay buffer       & $10^6$ \\
Quantiles per critic  & 25                   & Learning rate       & $3\times10^{-5}$ \\
Top quantiles dropped & 6                    & Optimizer           & AdamW \\
$\gamma$              & 0.97                 & Total env steps     & $10^6$ \\
$\tau$                & 0.005                & Warm-up steps       & 10\,000 \\
$\alpha$ (initial)    & 0.2 (auto)           & Updates per step    & 1 \\
Entropy target        & $-\dim(\mathcal{A})$ & Obs standardization   & running mean/std \\
\cmidrule(r){1-2}\cmidrule(l){3-4}
\multicolumn{2}{@{}l}{\textit{Networks}}  & Norm.\ freeze       & 150\,000 steps \\
Actor/Critic hidden   & {[}256, 256{]}       & \multicolumn{2}{l@{}}{\textit{Environment}} \\
Activation            & ReLU                 & Control frequency   & 1\,kHz \\
Auxiliary head        & linear               & PCS control rate    & 4\,kHz \\
Auxiliary loss weight & 1.0                  & Episode length      & 1\,s (1000 steps) \\
\cmidrule(r){1-2}
\multicolumn{2}{@{}l}{\textit{Dynamics randomization}} & Target resample     & 0.25\,s \\
Power supply voltage  & $\pm$50\,V (uniform) & Dropout prob.\ $p$  & 0.3 \\
\cmidrule(l){3-4}
                      &                      & \multicolumn{2}{l@{}}{\textit{Observation noise ($\sigma$)}} \\
                      &                      & Probes              & 0.01\,mT \\
                      &                      & Flux loops          & 0.01\,mWb \\
                      &                      & Coil currents       & 100\,A \\
\bottomrule
\end{tabular}
\end{table}

\begin{figure}[h]
  \begin{minipage}[t]{0.32\linewidth}
    \centering
    \includegraphics[width=\linewidth]{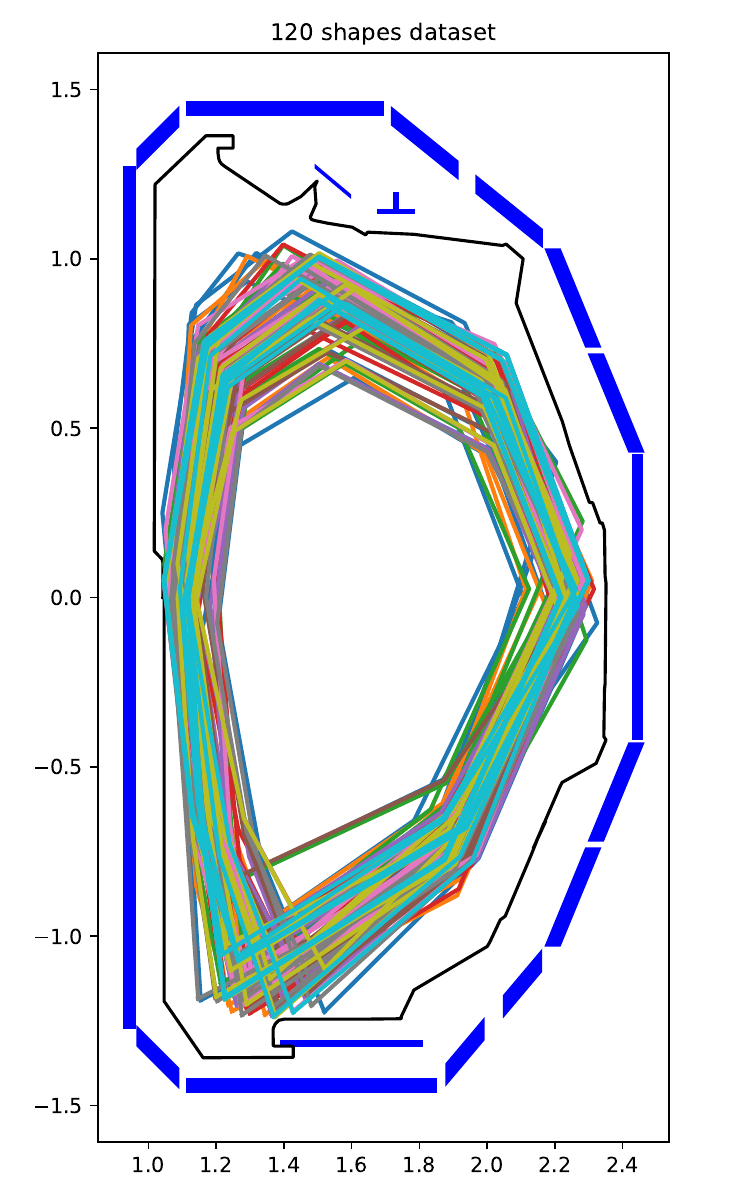}
    \caption{All 120 experimental LSN plasma boundaries used as the training distribution, overlaid on the DIII-D cross-section. Shapes span the full operational envelope including extreme elongations and x-point positions.}
    \label{fig:dataset}
  \end{minipage}\hfill
  \begin{minipage}[t]{0.64\linewidth}
    \centering
    \includegraphics[width=\linewidth]{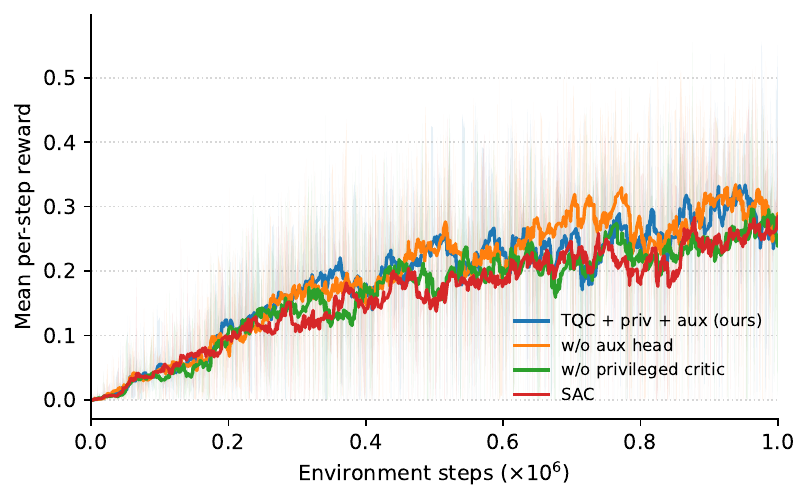}
    \caption{Training curves (mean per-step reward vs.\ environment steps) for the four ablation variants, all trained without diagnostic dropout. Shaded regions show per-episode reward; solid lines are exponential moving averages. The full model (TQC + priv + aux) and the no-aux variant reach similar final reward; removing the privileged critic or switching to SAC converges to a lower plateau.}
    \label{fig:training_curves}
  \end{minipage}
\end{figure}

\subsection*{Compute resources}

Each training run uses a single NVIDIA RTX A5000 GPU and takes approximately 24 hours ($10^6$ environment steps with one gradient update per step).
The reported experiments comprise 8 training runs — four dropout-rate sweeps ($p \in \{0.1, 0.3, 0.5, 0.7\}$) and four ablation variants — totalling approximately 8 GPU-days.
NSFsim evaluation is CPU-bound: evaluating a single agent across all 120 shapes takes 10--15 minutes using 30 CPU cores (roughly 2 minutes per shape).
The full research project involved several hundred preliminary training runs, representing substantially more compute than the reported experiments.

\section{Ablations and Sensor Importance Analysis}
\label{app:ablations}

\subsection{Component ablations}

To isolate the contribution of each architectural component from the confound of diagnostic dropout stochasticity, ablation variants are trained \emph{without} dropout and evaluated on the fixed DIII-D disabled-sensor mask across all 120 shapes; results are shown in Table~\ref{tab:ablations} and training curves in Figure~\ref{fig:training_curves}.

\begin{table}[h]
\centering
\caption{Component ablation results averaged over all 120 NSFsim shapes. All variants trained \emph{without} diagnostic dropout to isolate architectural effects; evaluated on the fixed DIII-D disabled-sensor mask. $\dshape$ and $\dxpt$ in cm; lower is better. Reward and episode length higher is better (max eplen = 250).}
\label{tab:ablations}
\begin{tabular}{lcccc}
\toprule
\textbf{Variant} & \textbf{Reward} & $\dshape$ \textbf{(cm)} & $\dxpt$ \textbf{(cm)} & \textbf{Eplen} \\
\midrule
TQC + priv + aux (full)  & $0.415 \pm 0.138$ & $4.0 \pm 3.1$ & $2.5 \pm 1.8$  & $249.9 \pm 0.7$  \\
TQC + priv, no aux       & $0.414 \pm 0.148$ & $4.8 \pm 5.5$ & $2.5 \pm 2.3$  & $247.0 \pm 21.0$ \\
TQC + aux, no priv       & $0.375 \pm 0.147$ & $4.9 \pm 3.9$ & $3.4 \pm 2.7$  & $248.3 \pm 12.0$ \\
SAC + priv + aux         & $0.362 \pm 0.138$ & $4.9 \pm 3.6$ & $3.9 \pm 16.7$ & $249.1 \pm 7.8$  \\
\bottomrule
\end{tabular}
\end{table}

\textbf{Asymmetric critic.}
Removing privileged equilibrium information causes the largest drop: reward falls from 0.415 to 0.375 and $\dxpt$ degrades from 2.5 to 3.4\,cm.
The actor structure is identical across variants, confirming that the benefit comes from better value estimates during training, not from architectural differences.

\textbf{Auxiliary shape loss.}
Removing the $\Delta\mathbf{p}$ prediction head has a negligible effect on reward and $\dxpt$, but raises $\dshape$ from 4.0 to 4.8\,cm and increases episode length std from 0.7 to 21.0 steps.
The aux loss appears to act as a training stabilizer, anchoring the actor's internal representation to a physically meaningful geometric quantity and preventing occasional early terminations.

\textbf{TQC vs.\ SAC.}
SAC achieves reward 0.362 vs.\ 0.415 for TQC, and shows a dramatically larger $\dxpt$ standard deviation (16.7\,cm vs.\ 1.8\,cm), indicating that it occasionally loses x-point control on harder shapes within the 120-shape distribution.

\begin{figure}[h]
  \centering
  \includegraphics[width=\linewidth]{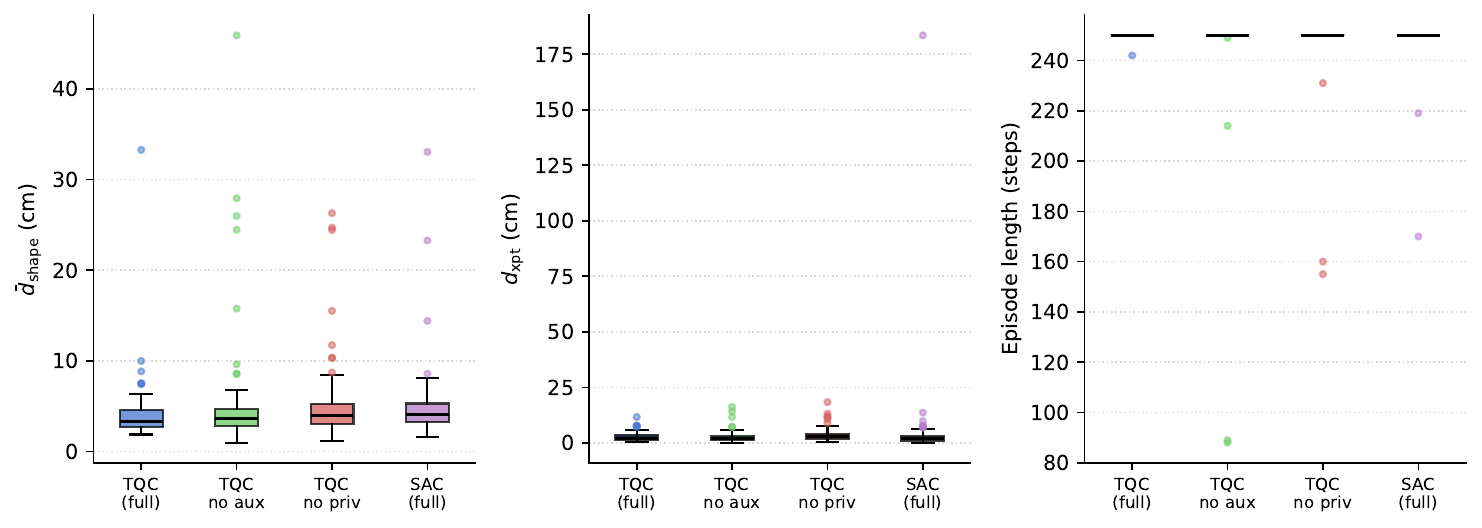}
  \caption{Per-shape distributions of $\dshape$ (left), $\dxpt$ (center), and episode length (right) across all 120 NSFsim shapes for four ablation variants. Boxes show the interquartile range; whiskers extend to $1.5\times$IQR; dots are outliers. The SAC $\dxpt$ outliers correspond to shapes where x-point control is lost.}
  \label{fig:ablation_boxes}
\end{figure}

Figure~\ref{fig:ablation_boxes} shows the full per-shape distributions, complementing the mean $\pm$ std summary in Table~\ref{tab:ablations}.
The SAC $\dxpt$ panel shows a small number of extreme outliers corresponding to shapes where the policy loses x-point control entirely; the TQC no-aux panel shows occasional early episode terminations visible in the eplen distribution.

\subsection{Sensor importance analysis}
\label{app:sensor_importance}

We analyze which magnetic diagnostics the policy relies on most using two complementary methods.

\textbf{Gradient sensitivity.}
For each sensor channel $i$, we compute the gradient of the auxiliary head prediction $\widehat{\Delta\mathbf{p}}$ with respect to the input $o_i$, averaged over a held-out set of NSFsim trajectories:
\begin{equation}
  s_i = \mathbb{E}\left[\left\|\frac{\partial \widehat{\Delta\mathbf{p}}}{\partial o_i}\right\|_2\right].
\end{equation}
The auxiliary head is preferred over the critic for this analysis because it predicts a physically interpretable quantity ($\Delta\mathbf{p}$, the pivot-point displacement), making gradient magnitudes directly meaningful.
Coil currents (${\sim}10^{-8}$--$10^{-9}$) and plasma current (${\sim}10^{-10}$) fall well below the displayed range and are omitted.

\begin{figure}[h]
  \hspace{-0.05\linewidth}%
  \begin{minipage}[t]{0.58\linewidth}
    \centering
    \includegraphics[height=0.52\textheight]{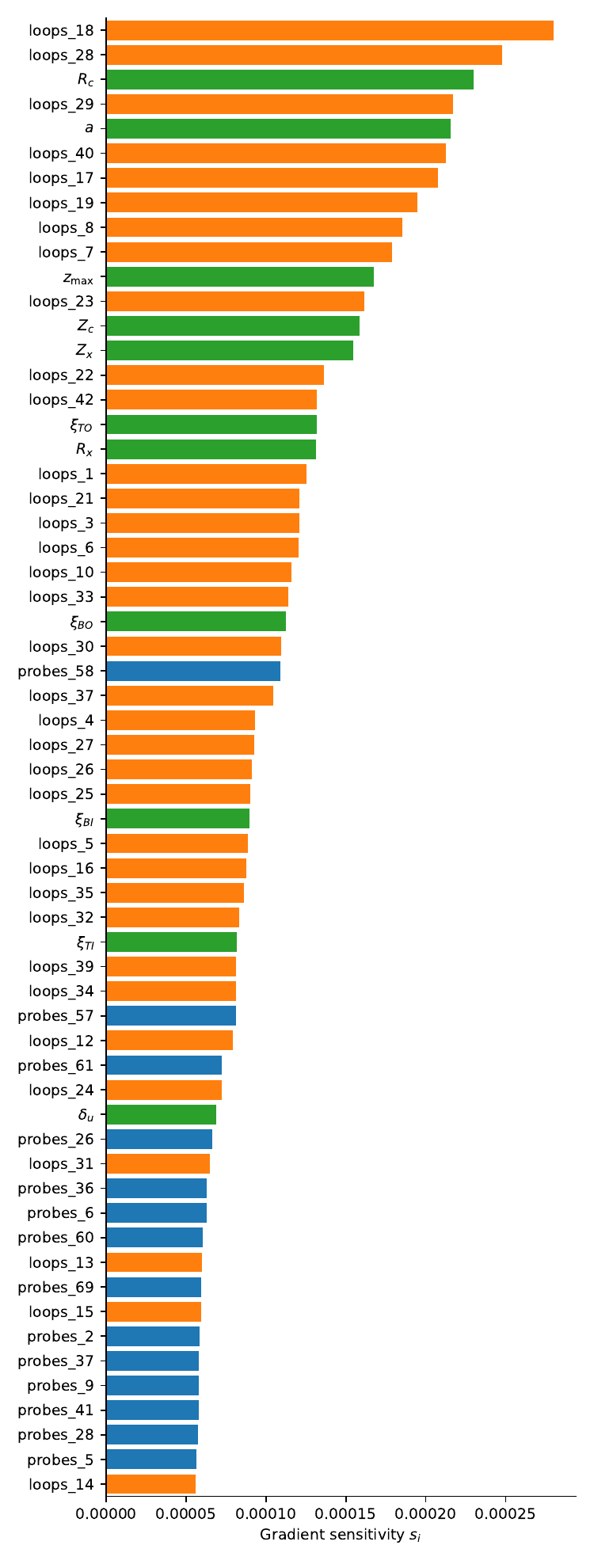}
  \end{minipage}\hspace{-0.05\linewidth}%
  \begin{minipage}[t]{0.38\linewidth}
    \centering
    \includegraphics[height=0.52\textheight]{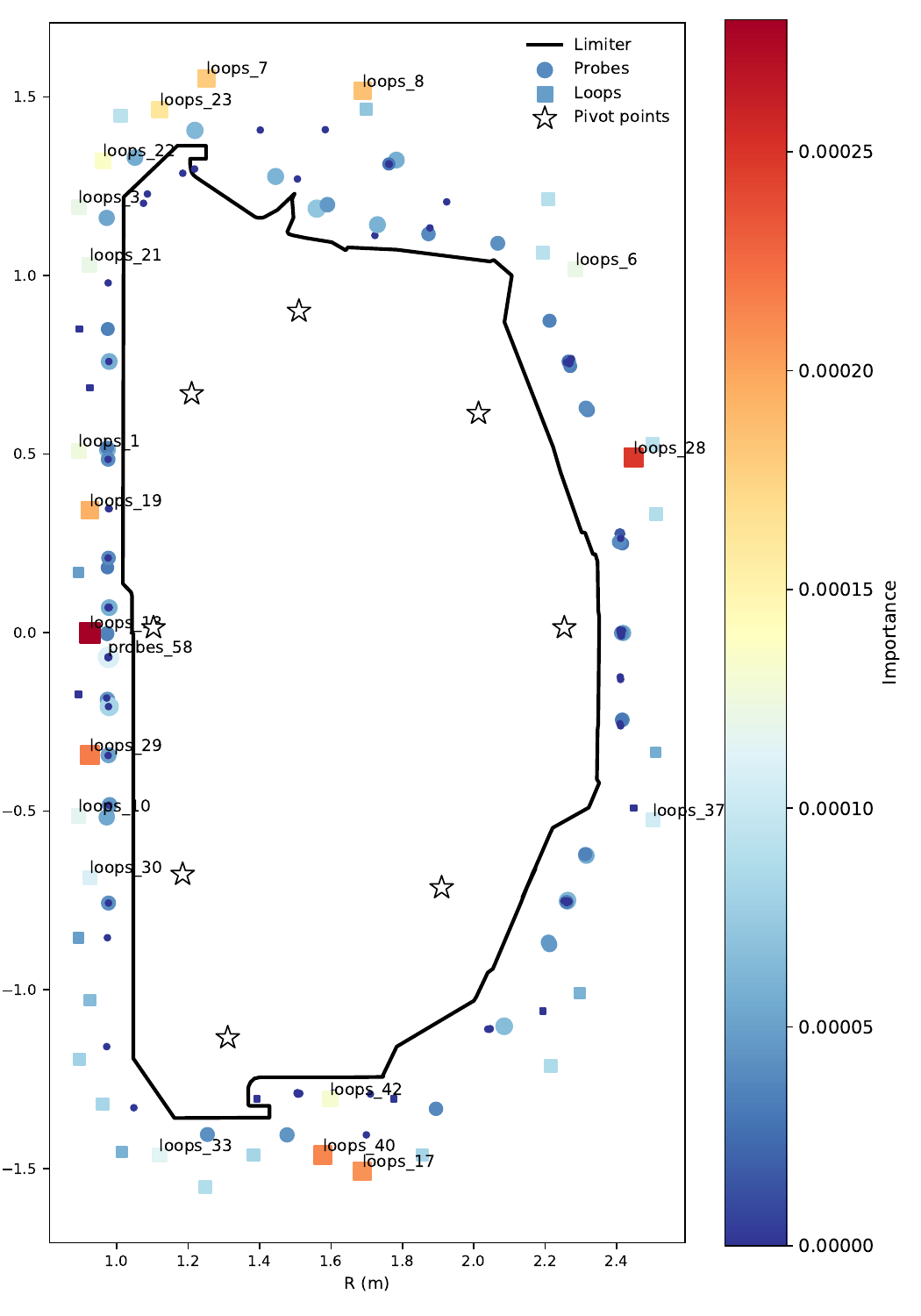}
  \end{minipage}
  \caption{Gradient sensitivity $s_i$ for the top 60 input channels, colored by sensor type: probes (blue), loops (orange), goals (green).
  Coil currents ($\sim$$10^{-8}$--$10^{-9}$) and plasma current ($\sim$$10^{-10}$) fall four or more orders of magnitude below the displayed range.
  \textit{Left:} channels sorted by $s_i$; all 11 goal components appear in the top tier alongside flux loops and probes, with $R_c$ ranking 3rd overall.
  \textit{Right:} spatial distribution of sensor importance overlaid on the DIII-D cross-section; dot size and color encode $s_i$; top-20 physical sensors are labeled by ID; white stars mark the 8 target pivot points; the highest-importance sensors cluster adjacent to the pivot points, with a notable concentration on the inner limiter side where the plasma--wall gap is smallest.
  All 11 goal parameters appear within the top 45 channels; the last to enter is upper triangularity $\delta_u$ (rank 45).}
  \label{fig:sensor_importance}
\end{figure}

Figure~\ref{fig:sensor_importance} shows the full ranking.
Among the non-zero channels, scores decay continuously from $2.8 \times 10^{-4}$ (loops\_18) to $1.6 \times 10^{-5}$ (probes\_46) with no sharp internal boundary.
Notably, goal channels rank among the most sensitive inputs: all 11 goal components appear in the top 60 alongside flux loops and probes, with $R_c$ reaching rank 3 overall.
The main structural break falls between the magnetic/goal group and coil currents: coil currents score $\sim$$10^{-8}$--$10^{-9}$, roughly four orders of magnitude below the top-ranked sensors, indicating the policy relies primarily on magnetic diagnostics and shape goals rather than actuator state for shape inference.
Plasma current $I_p$ falls a further order of magnitude below coil currents ($\sim$$10^{-10}$).
Spatially, the highest-importance physical sensors (Figure~\ref{fig:sensor_importance}, right) cluster near the 8 target pivot points (white stars) and are notably concentrated on the inner limiter side of the vessel, where the plasma--wall gap is smallest and local magnetic measurements carry the most information about plasma boundary position.

To assess whether the importance structure is specific to the deployment policy or a property of the task, we compute Spearman rank correlations between importance rankings from all four trained policies ($p \in \{0.1, 0.3, 0.5, 0.7\}$).
Across all 146 channels, pairwise correlations exceed 0.96, confirming that the coarse tier structure — which sensors matter vs.\ which do not — is stable across dropout rates.
Within the top 60 channels, correlations drop to 0.77--0.90, indicating that the fine-grained ordering within the high-importance tier is more sensitive to the training condition.

\begin{figure}[h]
  \centering
  \includegraphics[width=0.85\linewidth]{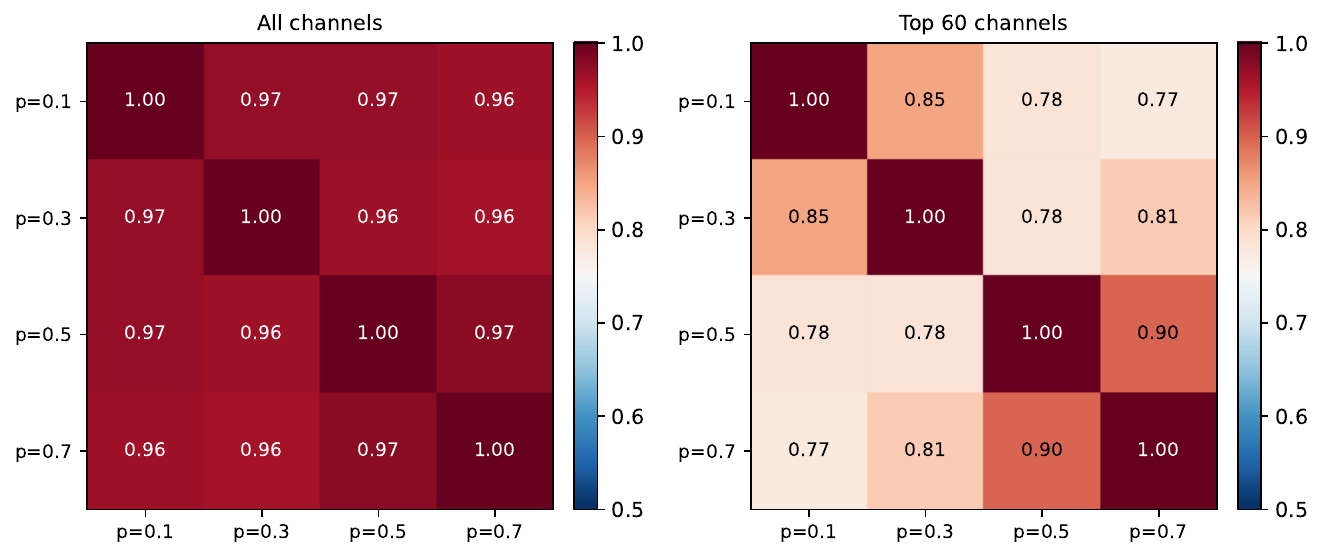}
  \caption{Pairwise Spearman rank correlations between gradient sensitivity rankings from policies trained at $p \in \{0.1, 0.3, 0.5, 0.7\}$.
  \textit{Left:} all 146 input channels ($\rho > 0.96$).
  \textit{Right:} top 60 channels by mean importance ($\rho = 0.77$--$0.90$).
  The coarse tier structure is stable across dropout rates; fine-grained ordering within the high-importance tier is more variable.}
  \label{fig:importance_corr}
\end{figure}

\textbf{Performance under top-$K$ sensors only.}
We evaluate all four trained policies ($p \in \{0.1, 0.3, 0.5, 0.7\}$) using their own gradient rankings.
For each $K$, we select the top $n\%$ of probes and the top $n\%$ of loops by gradient importance, where $n$ is chosen so that the total equals $K$.
This proportional selection ensures both sensor types are always represented, avoiding the degenerate case where a global top-$K$ ranking selects only loops at small $K$.
We sweep $K \in \{11, 23, 37, 58, 80, 114\}$ (corresponding to $10\%$, $20\%$, $33\%$, $51\%$, $70\%$, $100\%$ of each type) and compare against a random-$K$ baseline with the same proportional constraint, averaged over 5 draws per $K$.

\begin{figure}[h]
  \centering
  \begin{minipage}[t]{0.48\linewidth}
    \centering
    \includegraphics[width=\linewidth]{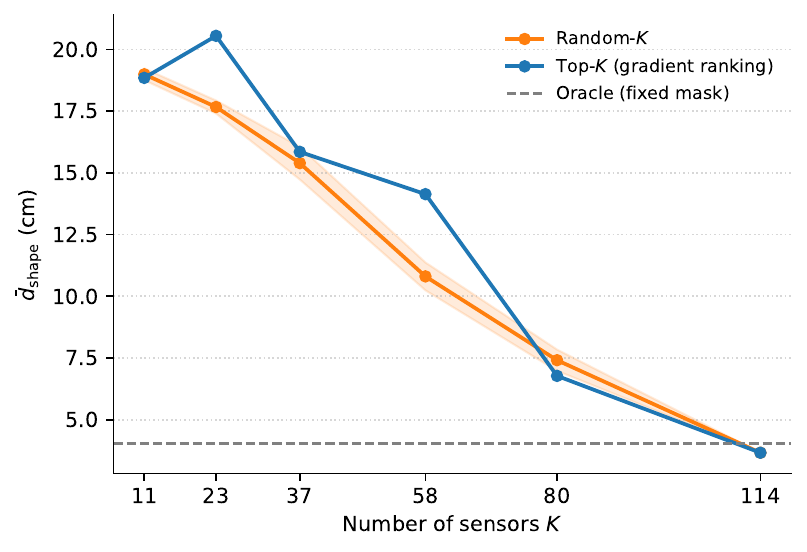}
    \subcaption{$p=0.1$}
  \end{minipage}\hfill
  \begin{minipage}[t]{0.48\linewidth}
    \centering
    \includegraphics[width=\linewidth]{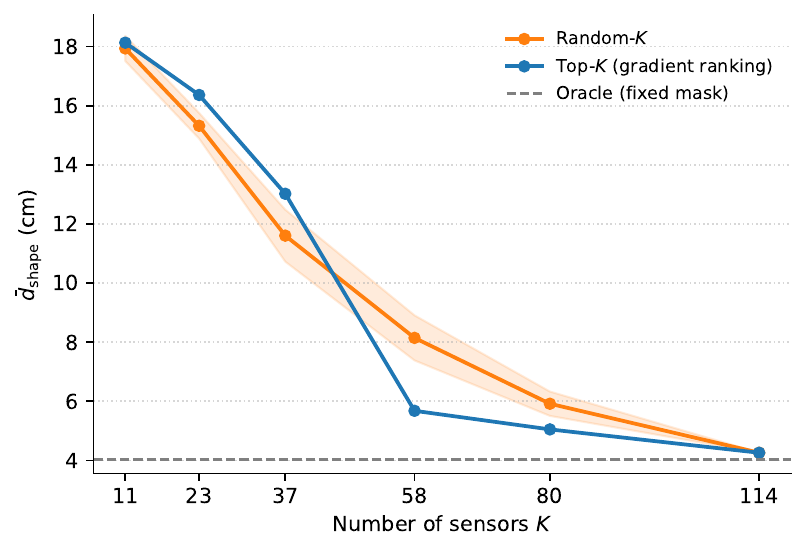}
    \subcaption{$p=0.3$}
  \end{minipage}
  \vspace{0.5em}
  \begin{minipage}[t]{0.48\linewidth}
    \centering
    \includegraphics[width=\linewidth]{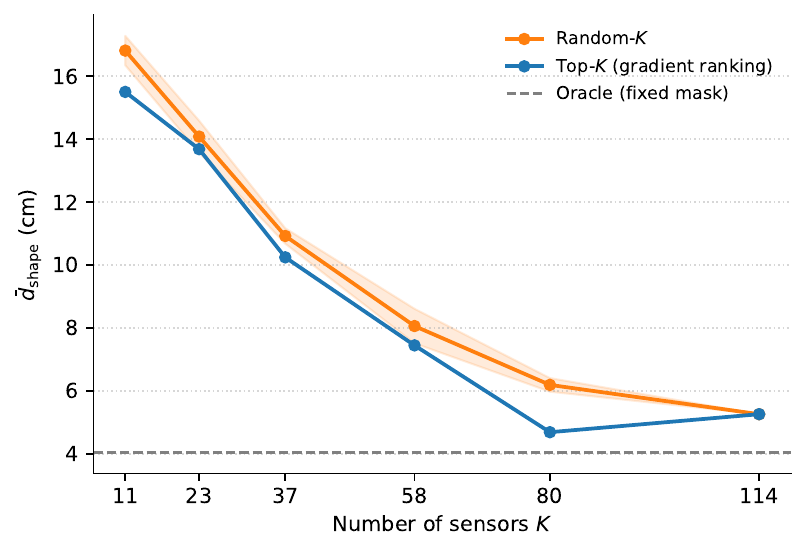}
    \subcaption{$p=0.5$}
  \end{minipage}\hfill
  \begin{minipage}[t]{0.48\linewidth}
    \centering
    \includegraphics[width=\linewidth]{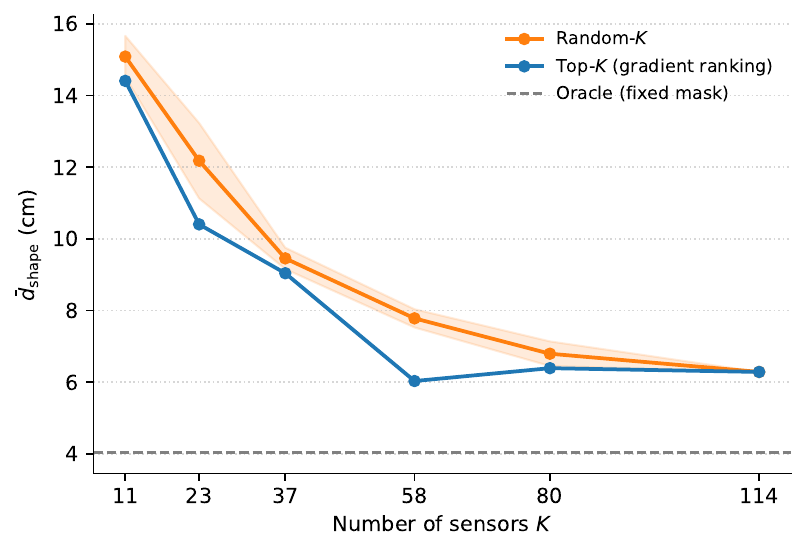}
    \subcaption{$p=0.7$}
  \end{minipage}
  \caption{$\dshape$ vs.\ number of available sensors $K$ for the top-$K$ gradient ranking (blue) and a proportionally random-$K$ baseline averaged over 5 draws (orange), for each dropout policy. Both conditions select the same percentage of probes and loops. Shaded bands show $\pm$1\,std. The dashed gray line marks the oracle model trained on the fixed DIII-D disabled-sensor mask. $K=114$ corresponds to all magnetic sensors active.}
  \label{fig:sensor_topk}
\end{figure}

Figure~\ref{fig:sensor_topk} shows the results.
For policies trained with higher dropout ($p=0.5$ and $p=0.7$), gradient ranking outperforms random selection at every $K$.
For the deployment policy $p=0.3$, gradient ranking provides substantial benefits at $K \geq 58$ ($\dshape = 5.7$ vs.\ $8.1$\,cm).
The $p=0.1$ policy is unsuitable for aggressive sensor reduction: trained with only 10\% dropout, it expects ${\approx}103$ active sensors and has rarely encountered configurations with fewer than ${\approx}80$ sensors during training.
At small $K$, the evaluation falls well outside this distribution, and no selection strategy can compensate for a policy that has not learned to operate under severe sensor loss.
Gradient ranking does induce a measurable separation from the random baseline at most $K$ values --- the gap exceeds the across-draw variance of the random-$K$ condition --- confirming the importance scores are non-trivial, but the underlying policy lacks the robustness to translate a better sensor subset into better control.
Non-monotonic behaviour (best performance below $K=114$) is observed for $p \geq 0.5$, consistent with low-importance sensors acting as noise beyond each policy's training distribution.

\subsection{Auxiliary head shape reconstruction on DIII-D shots}
\label{app:aux_reconstruction}

\begin{figure}[h]
  \centering
  \includegraphics[width=\linewidth]{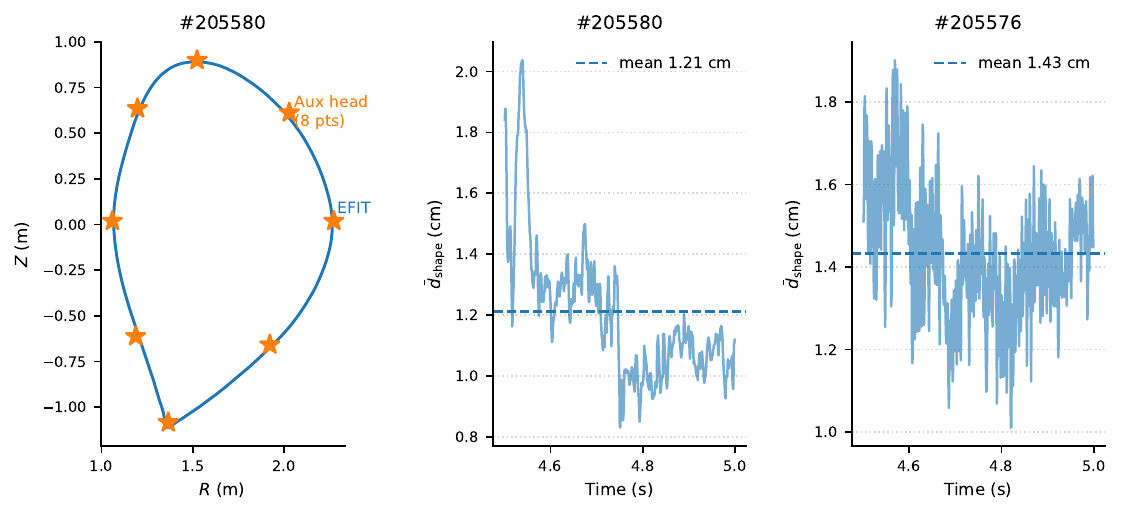}
  \caption{Auxiliary head vs.\ EFIT shape reconstruction on DIII-D shots.
  \textit{Left:} EFIT LCFS contour (blue) and the 8 reconstructed pivot points (orange) at a representative mid-episode timestep of Discharge \#205580; the pivot points lie on the EFIT boundary.
  \textit{Center/Right:} per-timestep mean distance from pivot points to EFIT for Discharges \#205580 and \#205576; dashed lines mark episode means of 1.21 and 1.43\,cm.
  The auxiliary head matches EFIT to within $1.2$--$1.4$\,cm despite operating at 4\,kHz on a partially masked diagnostic set.}
  \label{fig:efit_vs_aux}
\end{figure}

At inference the actor continuously estimates its own shape error from raw diagnostics via the auxiliary prediction head, with no separate reconstruction module required (§\ref{sec:algorithm}).
To validate this claim on the the physical device, we compare the pivot-point boundary reconstructed by the auxiliary head against the EFIT equilibrium reconstruction for the two experimental shots.
For each timestep we compute the mean distance from each of the 8 reconstructed pivot points to the nearest point on the EFIT LCFS contour.
The auxiliary head achieves $1.21 \pm 0.23$\,cm on Discharge \#205580 and $1.43 \pm 0.16$\,cm on Discharge \#205576, both within typical EFIT reconstruction uncertainty.
Figure~\ref{fig:efit_vs_aux} shows the reconstructed boundaries and the per-timestep distance trace for both shots.

\end{document}